\documentclass[10pt,twocolumn,letterpaper]{article}

\usepackage{wacv}              

\usepackage{graphicx}
\usepackage{amsmath}
\usepackage{amssymb}
\usepackage{booktabs}

\usepackage{amssymb}
\usepackage{mathrsfs}
\usepackage{graphicx}
\usepackage{multirow}
\usepackage{subcaption}
\usepackage{enumitem}
\usepackage{xcolor}
\usepackage{caption}
\usepackage[accsupp]{axessibility}
\usepackage[pagebackref,breaklinks,colorlinks,hyperfootnotes=false]{hyperref}
\usepackage{footnote}

\usepackage[capitalize]{cleveref}
\crefname{section}{Sec.}{Secs.}
\Crefname{section}{Section}{Sections}
\Crefname{table}{Table}{Tables}
\crefname{table}{Tab.}{Tabs.}

\def\wacvPaperID{523} 
\def\confName{WACV}
\def\confYear{2025}

\begin{document}
\title{Retrieval Augmented Recipe Generation}


\author{
Guoshan Liu\textsuperscript{1,2}\thanks{Equal contribution.}, Hailong Yin\textsuperscript{1,2}$^*$, 
Bin Zhu\textsuperscript{3}, 
Jingjing Chen\textsuperscript{1,2}\thanks{Jingjing Chen is the corresponding author.}, 
Chong-Wah Ngo\textsuperscript{3}, 
Yu-Gang Jiang\textsuperscript{1,2} \\
\textsuperscript{1}Shanghai Key Lab of Intelligent Information Processing, School of Computer Science, Fudan University \\
\textsuperscript{2}Shanghai Collaborative Innovation Center on Intelligent Visual Computing \\
\textsuperscript{3}Singapore Management University \\
{\tt\small \{gsliu24, hlyin23\}@m.fudan.edu.cn, \{chenjingjing, ygj\}@fudan.edu.cn} \\
{\tt\small \{binzhu, cwngo\}@smu.edu.sg} 
}


\maketitle

\begin{abstract}
  The growing interest in generating recipes from food images has drawn substantial research attention in recent years. Existing works for recipe generation primarily utilize a two-stage training method—first predicting ingredients from a food image and then generating instructions from both the image and ingredients. 
Large Multi-modal Models (LMMs), which have achieved notable success across a variety of vision and language tasks, shed light on generating both ingredients and instructions directly from images. 
Nevertheless, LMMs still face the common issue of hallucinations during recipe generation, leading to suboptimal performance. To tackle this issue, we propose a retrieval augmented large multimodal model for recipe generation. We first introduce Stochastic Diversified Retrieval Augmentation (SDRA) to retrieve recipes semantically related to the image from an existing datastore as a supplement, integrating them into the prompt to add diverse and rich context to the input image. 
Additionally, Self-Consistency Ensemble Voting mechanism is proposed to determine the most confident prediction recipes as the final output. It calculates the consistency among generated recipe candidates, which use different retrieval recipes as context for generation.
Extensive experiments validate the effectiveness of our proposed method, which demonstrates state-of-the-art (SOTA) performance in recipe generation on the Recipe1M dataset. 
\end{abstract}
\vspace{-0.2in}
\section{Introduction}
\vspace{-0.1in}
\label{sec:intro}

With the rising focus on food and health, food computing \cite{food_survey} has increasingly captured attention
and spurred various food related tasks, such as food recognition \cite{ciocca2016food, yang2010food, fvl1, food101sota, Food-101, vireosota, lgs, gyx, zhubin, sfz, jpk}, cross-modal recipe retrieval \cite{fvl2, Hierarchical, recipe1m, fvl4, Revamping, Deep-based}, recipe generation \cite{Inverse_Cooking, Structural_Representations, Structure-Aware, FIRE, majumder2019generating}, food recommendation \cite{Recipe_recommendation, freyne2010intelligent, elsweiler2017exploiting} and food logging \cite{sahoo2019foodai, cox2017food}. Previous research on food understanding has primarily focused on classifying food and ingredient recognition \cite{rodenas2022learning, liu2022transformer, zhang2022sequential, liu2020food}. However, due to the limited availability of detailed information on prepared foods, a comprehensive visual food recognition system should not only be able to identify the type of diet or its ingredients but also generate cooking instructions. Therefore, the task of recipe generation has become a significant task in the field of food computing.
\begin{figure*}[h]
  \centering
  \includegraphics[width=\linewidth]{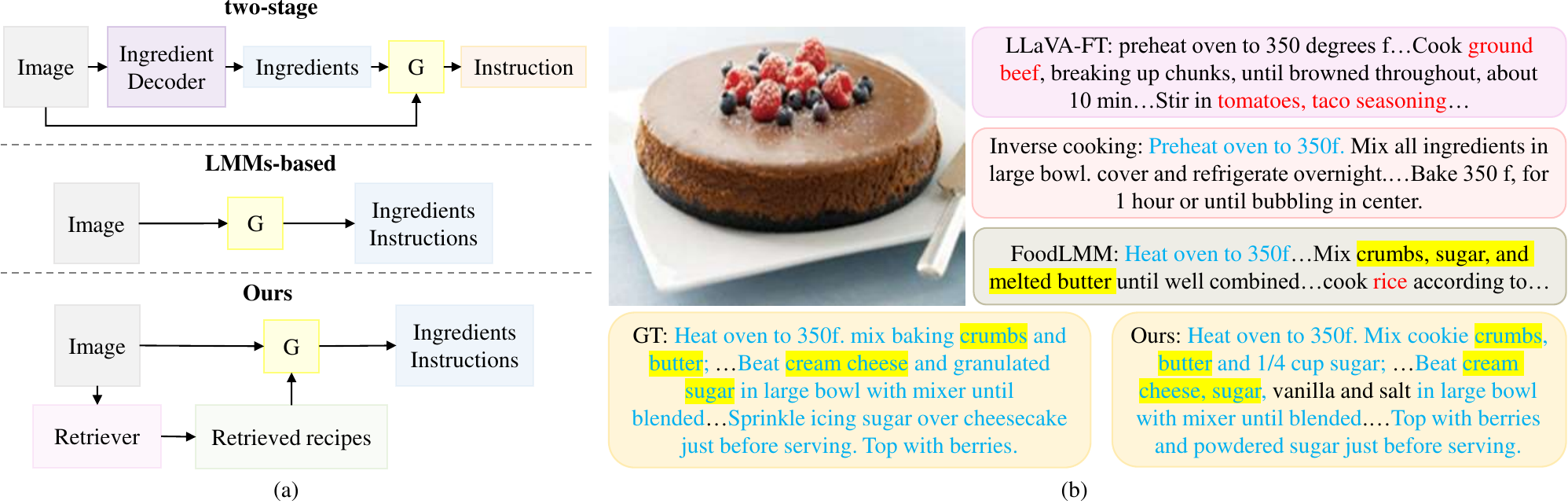}
  \captionsetup{aboveskip=0pt, belowskip=0pt}
  \caption{(a) The structural differences between our retrieval-augmented framework and the ``two-stage"~\cite{Inverse_Cooking, FIRE} and ``LMMs-based"~\cite{LLaVA, FoodLMM} approaches. ``G" refers to the generator. (b) Recipe generation results comparison. ``GT" refers to ground truth, ``LLaVA-FT" denotes the model using pre-trained LLaVA weights fine-tuned on Recipe1M, ``Inverse cooking" \cite{Inverse_Cooking} represents a model trained with two-stage, ``FoodLMM" \cite{FoodLMM} is the LMMs-based model for recipe generation, and ``Ours" refers to our model, where yellow highlights indicate ingredients that match those in the ``GT", blue signifies cooking instructions predictions matching the ``GT", and red font denotes incorrectly predicted ingredients.}
  \label{intro_case}
\vspace{-0.2in}
\end{figure*}
\vspace{-0.05in}

Previous methods for recipe generation \cite{Inverse_Cooking, DBLP:conf/wacv/MartinelFM18, Cross-modal} typically use a two-stage approach: first extracting ingredients from images, then generating instructions based on the embeddings of those ingredients and the images, which is shown in Figure~\ref{intro_case} (a). 
Due to limited training data and poor multi-modal alignment, traditional methods often yield unsatisfactory results. In contrast, Multi-modal Models (LMMs) \cite{gpt4, MiniGPT-v2, Instructblip, Flamingo, LLaVA} can directly generate recipes from images in one stage. FoodLMM \cite{FoodLMM} improved recipe generation performance but still suffers from hallucinations, affecting recipe quality.
Figure \ref{intro_case} (b) compares the recipe generation results of different methods. Compared to the ground truth instructions, the results predicted by one of the state-of-the-art LMMs, LLaVA \cite{LLaVA}, exhibit clear hallucinations, as it incorrectly identifies crumbs as `beef' and erroneously detects `tomatoes' and `taco seasoning' that are not present in the image. Although the two-stage method \cite{Inverse_Cooking} accurately identifies the correct temperature, it fails to precisely recognize the ingredients. FoodLMM \cite{FoodLMM} manages to identify most of the correct ingredients but still hallucinated, mistakenly recognizing `rice'. 
The result arises due to inadequate multi-modal understanding and a lack of effective use of context, which prevents the models from learning sufficient information. 

This paper addresses the limitation by introducing the first retrieval-augmented large multimodal model to generate recipes from food images. The proposed architecture consists of a retriever and a generator. The retriever leverages an off-the-shelf cross-modal recipe retrieval model \cite{Revamping} to identify semantically similar recipes from the image. The generator is built upon LLaVA \cite{LLaVA} with LoRA \cite{lora} to generate recipes based on the image and retrieved reference recipes. 
On the one hand, we propose Stochastic Diversified Retrieval Augmentation (SDRA) to provide a rich and diverse set of retrieved recipes as references, which could provide relevant knowledge for the generation to reduce hallucination~\cite{In-Context, REALM, Active_Retrieval_Augmented_Generation}. Section 2 of the supplementary materials compares retrieved recipes with the ground truth, highlighting related content that helps alleviate hallucinations and improve generation. On the other hand, we propose Self-consistency Ensemble Voting strategy to improve generation quality during the inference phase. Specifically, the generator produces multiple recipe candidates using different retrieved recipes and the image. Cosine similarity scores are computed for these candidates to assess their mutual agreement. The recipe with the highest score relative to the others is selected as the final output. The consensus among different candidates is capable of further reducing the hallucination in the generated recipes, as detailed in Section 5 of the supplementary materials, which explains why the final output is superior. On Recipe1M dataset \cite{recipe1m}, our proposed model significantly outperforms existing methods and fine-tuned LLaVA. Moreover, our model demonstrates strong generalizability, surpassing the state-of-the-art (SOTA) benchmarks on the Recipe1M dataset for ingredient recognition metrics.

Overall, our main contributions can be summarized as follows:
\begin{itemize}[itemsep=0pt]
\item We propose the first retrieval-augmented large multimodal model tailored for recipe generation. Our model introduces a stochastic diversified retrieval-augmentation technique to enhance the diversity and richness of retrieval. Additionally, we employ a Self-consistency Ensemble Voting strategy during the inference stage, using different retrieved recipes to ensure agreement and consistency in the generated recipe.
\item Our proposed model achieves SOTA recipe generation performance on Recipe1M dataset. We conduct comprehensive ablation studies to validate the effectiveness of our design choices and demonstrate the contributions of each component to the overall performance. Our proposed model exhibits exceptional adaptability, outperforming current SOTA results in ingredient recognition on the Recipe1M dataset.
\end{itemize}
\vspace{-0.1in}
\section{Related work}
\subsection{Recipe Generation}
The significance of food and the accessibility of comprehensive food datasets, including Recipe1M \cite{recipe1m}, Vireo Food-172 \cite{Deep-based} and Food2K \cite{Food2K}, have facilitated computational studies in the domain of food-related computing tasks \cite{food_survey}. Recipe generation poses a significant challenge due to the presence of multiple sentences in cooking instructions. It entails the complex task of generating food recipes based on provided food images \cite{Recipegpt, Structural_Representations, Structure-Aware}. 
Accurate recipe generation requires understanding food components, images, and processes. Early methods generated ingredients from images first, then instructions from these ingredients and images. \cite{Inverse_Cooking} used transformers for recipe generation but missed some steps due to a lack of comprehensive structure.
Previous efforts did not use Large Multi-modal Models (LMMs). Recently, \cite{FoodLMM} fine-tuned the LMM LISA \cite{lisa} using multiple datasets, creating the first unified food computing model, including recipe generation. However, FoodLMM still suffers from hallucination issues. This paper aims to mitigate these issues with retrieval augmentation.
\vspace*{-0.1in}
\subsection{Vision-language Multimodal Models}
\vspace*{-0.1in}
Due to the increasing demand for versatile deep learning models, various large pre-trained models like BERT \cite{BERT}, ViT \cite{VIT}, and GPT \cite{GPT} have emerged. 
However, their single-modality limits generalization, leading to the development of multimodal models.
Autoregressive language models are now popular for vision-language tasks \cite{gpt4, Flamingo, VisualGPT, BLIP-2, BLIP, jiao2024lumen, li2024eyes, zhang2024eventhallusion, zhang2024eagle}. For example, LLaVA \cite{LLaVA} integrates visual encoder output with LLaMA \cite{LLAMA} using synthetic data, while Vicuna \cite{vicuna} uses LLaMA for conversational interactions. The rise of LMMs has expanded their application in various domains. However, hallucination remains an issue. \cite{Evaluating} treats it as a binary classification problem, and \cite{FIGURE} uses models to generate data for annotators to identify hallucinations.
Our model is built upon LLaVA for recipe generation, equipped with a diversified stochastic retrieval augmentation, that boosts the recipe generation capabilities by introducing additional relevant contextual information, enabling the model to learn more specialized and comprehensive information. 
\vspace*{-0.1in}
\subsection{Retrieval-Augmented Generation}
\vspace*{-0.1in}
Retrieval-augmented generation (RAG) enhances language models (LMs) by incorporating knowledge from an external datastore \cite{DBLP:conf/nips/LewisPPPKGKLYR020}. This involves fetching relevant documents from external storage to improve the LM's predictive accuracy \cite{Active_Retrieval_Augmented_Generation, In-Context, Knowledge-Intensive, REALM, chen2024benchmarking, gao2023retrieval}. While RAG is popular in NLP tasks \cite{DBLP:conf/nips/LewisGGAWZ20}, it is less explored in multimodal models \cite{Multi-Modal_Classification, jiao2021two}. Some relevant works in retrieval-augmented multimodal language modeling focus on caption generation based on the encoded input image, as well as a collection of retrieved texts which are used as a task demonstration, input to the decoder as a prompt \cite{smallcap}.
Recent works \cite{Retrieval-Augmented_Multimodal, REVEAL} train generators with external multimodal information. While RAG enhances language models by incorporating external data, its use in vertical domains is limited \cite{X-TRA, Llava-med}. We propose the first Retrieval-Augmented LMMs for recipe generation using a Stochastic Diversified Retrieval Augmentation (SDRA) method. This method employs a pretrained retriever to fetch diverse recipes as supplemental inputs, improving LMM capabilities in recipe generation.
\begin{figure}[!t]
  \centering
  \includegraphics[width=\linewidth]{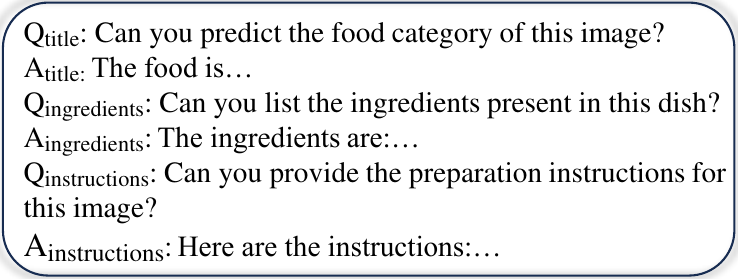}
  \captionsetup{skip=1pt}
  \caption{Templates for Recipe Generation.}
  \label{templets}
\vspace{-0.2in}
\end{figure}
\begin{figure*}[h]
  \centering
  \includegraphics[width=\linewidth]{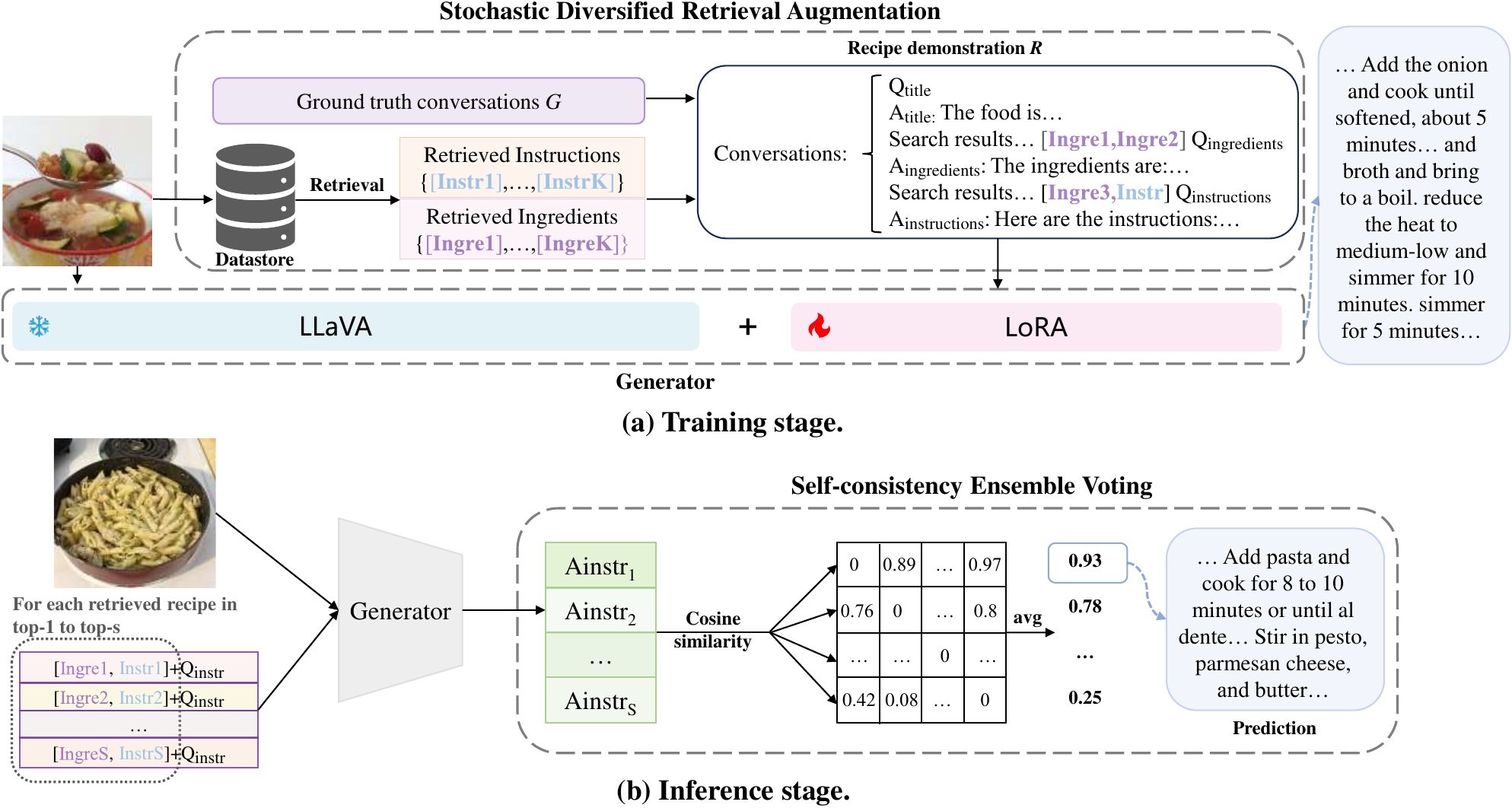}
  \captionsetup{aboveskip=0pt, belowskip=0pt}
  \caption{Overview of our proposed model architecture. Our model consists of a retriever to search semantically similar recipes from the image as reference, and a generator based on a frozen LLaVA \cite{LLaVA} with a trainable LoRA \cite{lora} to generate recipe with the image and retrieved recipes. \textbf{Stochastic Diversified Retrieval Augmentation} is introduced by using retrieved ingredients and instructions, 
   to form Recipe demonstration $R$, and fed into the generator for training. \textbf{Self-consistency Ensemble Voting} is proposed to select the final recipe output based on mutual agreement among the recipe candidates, which are produced by using each recipe from top 1 to top s retrieved recipes as context.}
  \label{framework}
\vspace{-0.2in}
\end{figure*}
\vspace{-0.1in}
\section{Method}  \label{all_method}
\subsection{Preliminary} \label{method}
\subsubsection{Retrieval Augmented Generation (RAG) Models}
\hspace*{1em} RAG models \cite{In-Context, Retrieval-Augmented_Multimodal, REALM} are usually composed of a retrieval model $R$ and a generator (language model) $G$. The retriever $R$, based on the query (input sequence) $x_1,...,x_n$, vectorizes the query to tokens and searches the Top $K$ documents with the highest similarity to the query sequence within the datastore M, denoted as  $M'=(m_1,...,m_K)$.
$M'$ is then concatenated with the query to form a context-rich prompt, which is fed into the language model to generate enhanced outputs. 
Next-token prediction is widely adopted in language models.
During the training process, the autoregressive model aims to maximize the conditional probability of the next word given the preceding sequence of words, namely, by optimizing the parameters $\theta$ to maximize: 
\setlength{\abovedisplayskip}{5pt}
\setlength{\belowdisplayskip}{5pt}
\begin{equation}
    p(x_1,...,x_n)=\prod \limits_{i=1}^{n}p_{\theta}(x_i|x_{<i}),
\end{equation}
where $x_{<i}$ is the sequence of tokens preceding $x_i$. Retrieval-augmented language models make predictions conditionally based on the retrieved documents $M'$. Specifically, by simply merging the retrieved documents into the input query to predict the continuation of the input:
\begin{equation}
    p(x_1,...,x_n)=\prod \limits_{i=1}^{n}p_{\theta}(x_i|[\mathcal{M'}(x_{<i});x_{<i}]),
\end{equation}
where $[a; b]$ denotes the concatenation of strings $a$ and $b$.
\vspace{-0.1in}
\subsubsection{Data Organization Strategy} 
\vspace*{-0.1in}

\hspace*{1em}We organize image-recipe pairs as dialogues, with each image linked to three question-answer pairs (titles, ingredients, instructions). In the retrieval model, the image $x$ is the query; in the generator, questions about titles, ingredients, and instructions serve as queries. During the training process, the model formats each multimodal document as ``[$<$Image$>$ Conversations: {$Q_{titles}$, $A_{titles}$, $Q_{ingredients}$, $A_{ingredients}$, $Q_{instructions}$, $A_{instructions}$}]", which $Q_{titles}$ denotes the title's questions and $A_{titles}$ denotes the title's answers of this image, and similarly for ingredients and instructions. We define these dialogues as ground truth conversations $G$. See Figure \ref{templets} for more details of the templates for our task.
\subsection{Retrieval Augmented Recipe Generation} \label{Retrieval strategy}
\subsubsection{Stochastic Diversified Retrieval Augmentation} 
\hspace*{1em}As depicted in Figure \ref{framework} (a), we utilize image-to-recipe retrieval model \cite{Revamping}, to retrieve the top K most similar ingredients and instructions from the data storage to the image $x$. 
Unlike existing retrieval augmentation methods \cite{smallcap, DBLP:conf/nips/LewisPPPKGKLYR020, Active_Retrieval_Augmented_Generation} which directly use the retrieved results as context, to ensure the diversity of retrieval information, we randomly sample from the top K retrieval results, selecting three sets of ingredients and one set of instructions as the final retrieval information, as shown in Figure \ref{framework} (a). 
1 set of ingredients refers to the ingredients part retrieved from an image, specifically, such as `oil, egg, milk, vanilla'.
Two sets of retrieved ingredients are concatenated in sequence before $Q_{ingredients}$, and the instructions and remaining set of retrieved ingredients are concatenated in sequence before $Q_{instructions}$ while informing the model like this that it is a reference result: ``Search results for reference is `retrieved information'. The search results are only for referring, please focus on the image." Finally, they are normalized as recipe demonstration $R$ according to the prompt in Figure \ref{framework} (a). Here, the prompts ``The food is", ``The ingredients are:" and ``Here are the instructions:" are similar to the simple, fixed prompts used in other research \cite{BLIP}.
\vspace*{-0.1in}
\subsubsection{Recipe Generation with Retrieval Recipes} 
\vspace*{-0.1in}
\hspace*{1em}Our generation model is built upon a Large Multi-modal Model LLaVA \cite{LLaVA} which takes image and text prompts as input.
The text prompt (i.e., questions) and the retrieved recipes are concatenated and processed through a tokenizer, then fed into a text encoder to obtain textual features, while the image is processed through an image encoder to obtain image features. The image features are further mapped into the same embedding space as text features via a MLP.
Upon receiving both text and image embeddings, the decoder proceeds to produce caption tokens, which are contingent on the image features $X$ and the recipe demonstration $R$. To reduce the compute requirements for training and to preserve their generalization capabilities, we freeze the generator LLaVA model and only train its patch LoRA \cite{lora}, which allows the focus to then be on fine-tuning specific, smaller aspects of the model to adapt to recipe generation without the need for extensive computation. 
The model is trained by minimizing the cross-entropy loss for next token prediction as follows:
\begin{equation} \label{loss}
    L_\theta=-\sum \limits_{i=1} ^{n}p_{\theta}(y_i|[X;R;y_{<i}];\theta),
\end{equation}
where $n$ is the index of the current tokens, and $y_{<i}$ denotes represents the tokens in the sequence before position $i$.
\vspace{-0.1in}
\subsection{Self-consistency Ensemble Voting}
\vspace*{-0.1in}
To further improve the quality of recipe generation, we propose self-consistency ensemble voting.
As shown in Figure \ref{framework} (b), we first retrieve the Top S sets of ingredients and instructions from the food image during the testing phase. Each set is then concatenated before $Q_{instructions}$ in the model input, following the Recipe denomination $R$ in Figure \ref{framework} (a), which concatenates the retrieved data in front of $Q_{instructions}$. The S retrieved sets are sequentially concatenated before $Q_{instructions}$ and input into the model, producing S different outputs. This process generates S different recipes, we note that the generated recipes could be inconsistent with different retrieved recipes as context.
As a result, we introduce a score-based ensemble voting method, which selects the best recipe from multiple predictions to maintain self-consistency and improve the quality of the generated recipes.
Specifically, denote the generated recipes as $P=\{P_1,...,P_S\}$ by employing top 1 to top $S$ as retrieved recipes, where $S$ is the number of recipes used for inference.
We compute the cosine similarity, BLEU, SacreBLEU scores, or ROUGE L scores among these $S$ recipes, producing a $S\times{S}$ matrix where each row represents the agreement of all other predictions with the current prediction (excluding diagonal elements). By averaging the sum of agreements in each row, we obtain a confidence score for each recipe. Then, taking the calculation of confidence scores using cosine similarity as an example, we select the recipe with the highest confidence score as our final output as follows:
\setlength{\abovedisplayskip}{5pt}
\setlength{\belowdisplayskip}{5pt}
\begin{equation} 
    P_{\text{best}} = \arg \max_i \left( \frac{1}{S-1} \sum_{j=1, j \neq i}^{S} \frac{P_i \cdot P_j}{\|P_i\| \|P_j\|} \right),
\end{equation}
where $\frac{P_i \cdot P_j}{\|P_i\| \|P_j\|}$ refers to cosine similarity between two predictions.
\vspace{-0.1in}
\begin{figure*}[h]
  \centering
  \includegraphics[width=\linewidth]{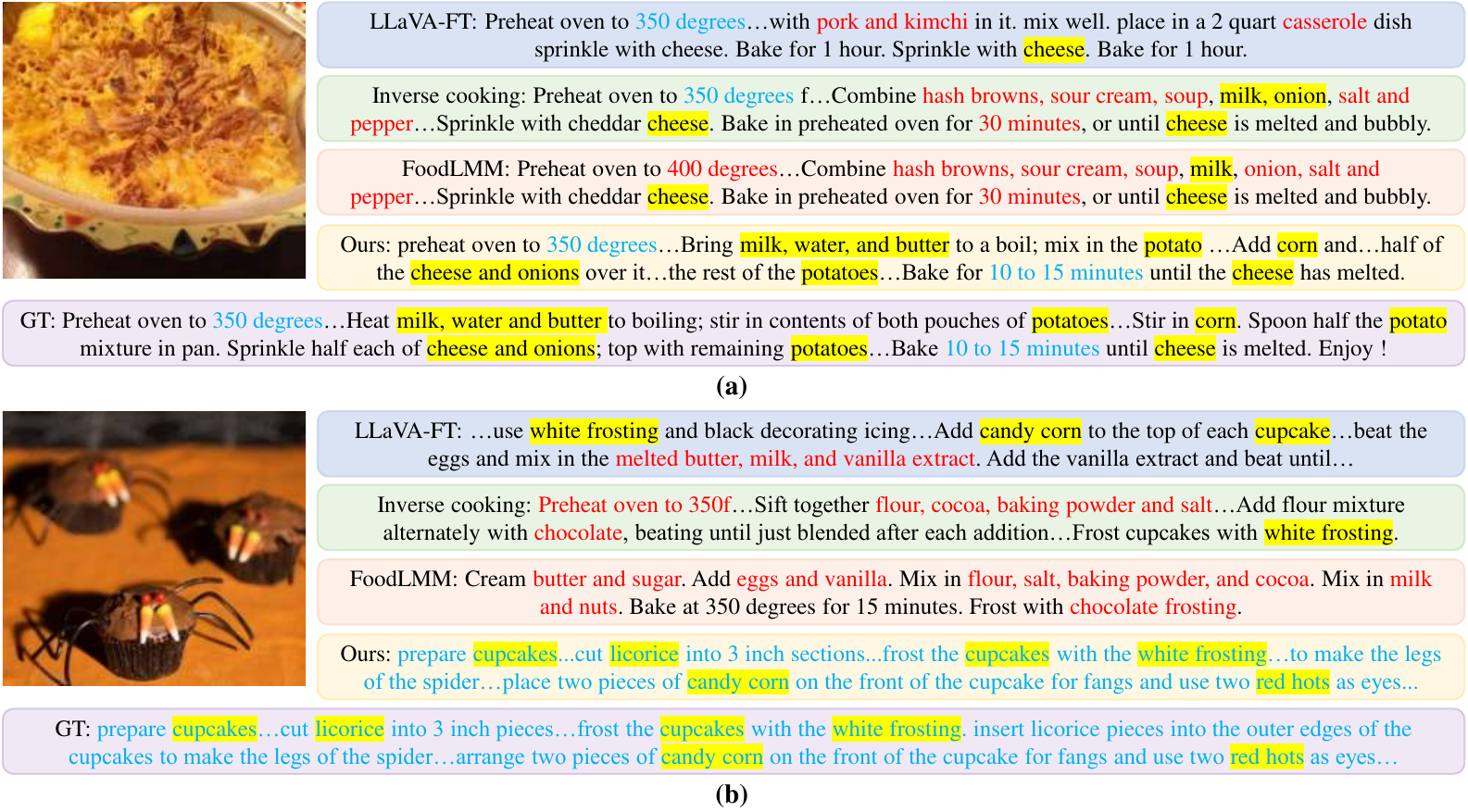}
  \captionsetup{skip=1pt}
  \caption{Qualitative results. The ingredients in generated recipes that overlap with ground truth (``GT") are highlighted in yellow, while details in the instructions that match the GT are shown in blue. Otherwise, the incorrect generation results are displayed in red. Best viewed in color.}
  \label{quality}
\vspace{-0.2in}
\end{figure*}
\vspace*{-0.1in}
\section{Experiments}
\subsection{Experimental Settings} \label{1488}
\subsubsection{Dataset and Evaluation Metrics} 
\vspace*{-0.1in}
\hspace*{1em}Following prior works \cite{FIRE, FoodLMM}, we use the recipes with images in Recipe1M dataset \cite{recipe1m} for experiments, including 571,587 pairs for training and 51,304 for testing. In this dataset, each image is associated with a title, ingredients, and instructions. During training, each image corresponds to three question-answer pairs as mentioned in Section \ref{method}.
Ingredients are clustered into 1,488 categories, and quantifiers are removed, in line with \cite{Inverse_Cooking}. For testing, the first image of each recipe is used, formatted to one question-answer pair per image, to separately test titles, ingredients, and instructions.
Generating long recipes word by word is time-consuming, so we randomly select 5,000 samples from test set and fix the seed for all experiments. Similar to \cite{Retrieval-Augmented_Multimodal}, we use the Recipe1M training set as our external datastore $M$ to ensure consistency and fairness, avoiding external data.

Following existing works \cite{Structural_Representations, FIRE}, we utilize F1 score and Intersection Over Union (IOU) to evaluate the quality of ingredients, as well as document-level evaluation metrics, specifically BLEU, SacreBLEU, and RougeL, to evaluate the quality of instructions in the generated recipes.

\vspace{-0.1in}
\subsubsection{Implementation} \label{implementation}
\vspace*{-0.1in}
\hspace*{1em}In our retrieval module $R$, we use the revamping cross-modal recipe retrieval model \cite{Revamping} based on Transformers. Our generator $G$ utilizes LLaVA \cite{LLaVA} augmented with a LoRA \cite{lora} patch. During training, we utilize LoRA to fine-tune LLaVa based on the PyTorch framework, employing the pre-trained weights from LISA-7B-v1-explanatory \cite{lisa}. This process is carried out on four NVIDIA 80G A100 GPUs. 
For the baseline, which we named ``LLaVA-FT," we fine-tuned LLaVA using LoRA without retrieval augmentation, utilizing the same model architecture, training data, and computational resources to ensure a fair comparison.

\subsubsection{Training and Inference} \label{inference} 
\vspace*{-0.1in}
\hspace*{1em}During training, we retrieve the top 50 recipes for each image, including titles, ingredients, and instructions. From these, we randomly select three sets of ingredients and one set of instructions to concatenate before $Q_{ingredients}$ and $Q_{instructions}$ as described in Section \ref{Retrieval strategy}.
The generator's max sequence length of 4096 encompasses all the information. We optimize the token prediction loss over the entire sequence (Equation \ref{loss}). Given the strong performance of our cross-modal recipe retriever \cite{Revamping}, which uses a transformer to encode recipe components, we keep the retriever constant and focus on training the generator. Future research could explore co-training or fine-tuning the retriever.

During inference, we first use the images from the test set as queries to retrieve from the 571,587 instances in the train set, with the retrieval model and method as mentioned in Section \ref{Retrieval strategy}. Then, using the retrieved instructions and ingredients, we concatenate them in front of $Q_{instructions}$ in the same pattern as during training. For Self-consistency Ensemble Voting, we sequentially use the $top s=\{top 1,...,top s\}$ retrieved information, adding it before $Q_{instructions}$ and then feeding the recipe demonstration $R$ in the form of ``[$<$Image$>$ Conversations: {Search results...[Ingre, Instr] $Q_{instructions}$, $A_{instructions}$}]" into the generator to obtain $s$ prediction results. The final prediction is made using the Self-consistency Ensemble Voting method. 
\begin{figure}[!t]
  \centering
  \includegraphics[width=\linewidth]{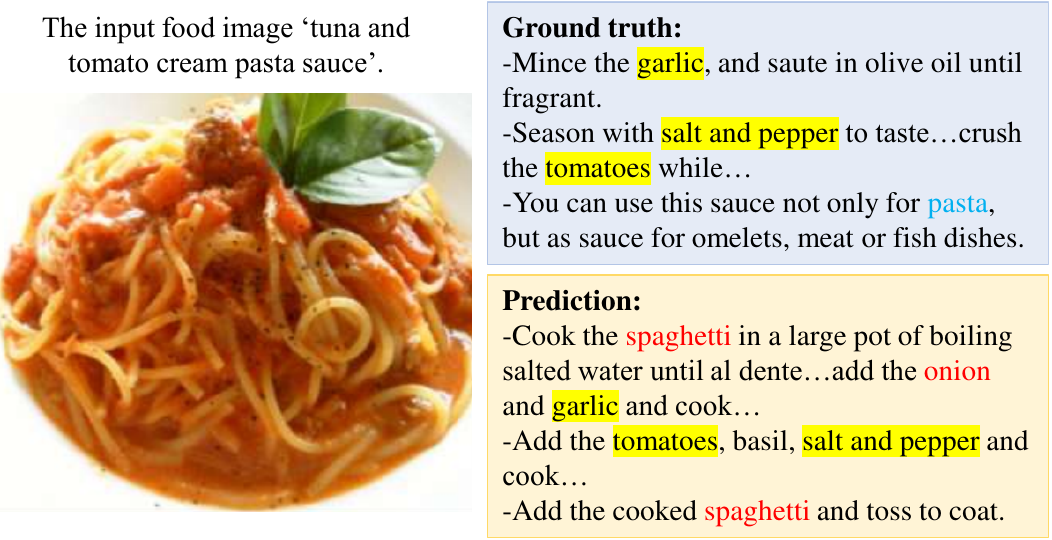}
  \caption{Comparison between generated recipes and GT recipes. The highlights in yellow indicate ingredients that match those in the GT, ingredients incorrectly identified by the model are signified in red.}
  \label{bad_case}
\vspace{-0.2in}
\end{figure}
\vspace{-0.1in}
\begin{table}[]
\centering
\setlength{\abovecaptionskip}{1pt}
\caption{Recipe generation performance comparison. ``LLaVA-FT" refers to fine-tuned LLaVA model.}
\label{sota}
\small
\begin{tabular}{c|ccc}
\hline
\textbf{Methods}        & \textbf{BLEU} & \textbf{SacreBLEU} & \textbf{ROUGE L} \\ \hline
Chef Transformer \cite{T5}        & 18.08             & 4.61               & 17.54            \\ \hline
InverseCooking \cite{Inverse_Cooking}          & 7.23          & 5.48               & 19.47            \\ \hline
TIRG \cite{TIRG}                    & 7.95          & —                  & 32.4             \\ \hline
VAL \cite{VAL}                     & 8.83          & —                  & 34.20             \\ \hline
SGN \cite{Structural_Representations}                     & 12.75         & —                  & 36.90             \\ \hline
FIRE \cite{FIRE}                    & —             & 6.02               & 21.29            \\ \hline \hline
FoodLMM \cite{FoodLMM}                    & 27.86             & 6.24               & 36.96            \\ \hline
LLaVA-FT \cite{LLaVA}      &28.32               &5.88                    &38.18                  \\ \hline
\textbf{Ours} &\textbf{30.11}               &\textbf{6.42}                    &\textbf{38.93}                  \\ \hline
\end{tabular}
\vspace{-0.2in}
\end{table}
\subsection{Performance Comparison}

\subsubsection{Quantitative Comparison of Recipe Generation} 
\vspace*{-0.1in}
\hspace*{1em}Table \ref{sota} demonstrates that our proposed method outperforms all the existing works by a noticeable margin. In particular, our method manages to surpass the recent LMM-based models, including LLaVA-FT (fine-tuned LLaVA)~\cite{LLaVA} and FoodLMM~\cite{FoodLMM}. We achieve a relative improvement of 2.25\%, 0.18\%, and 1.97\% over FoodLMM in BLEU, SacreBLEU and RougeL scores, respectively. These results demonstrate that our proposed framework can generate more precise and coherent recipes, confirming the effectiveness of our model.
\vspace{-0.1in}
\subsubsection{Quantitative Comparison of Ingredients Recognition} 
\vspace*{-0.1in}
\hspace*{1em}For the inference of ingredients, the input for each image is $Q_{ingredients}$, which directly generates the answers for the ingredients. Table \ref{recipe1m} lists the performance of ingredient recognition with existing methods in Recipe1M dataset. Our proposed method is superior to all the methods, with 1.05\% and 1.03\% improvement in terms of both F1 and IOU respectively, compared to FIRE~\cite{FIRE}. Note that FIRE and InverseCooking both specifically design an ingredient recognition network. In contrast, our method is capable of generating the ingredients and instructions in a conversational manner. 
\vspace{-0.1in}
\begin{figure*}[ht!]
  \centering
  \begin{subfigure}{0.32\textwidth}
    \includegraphics[width=\linewidth]{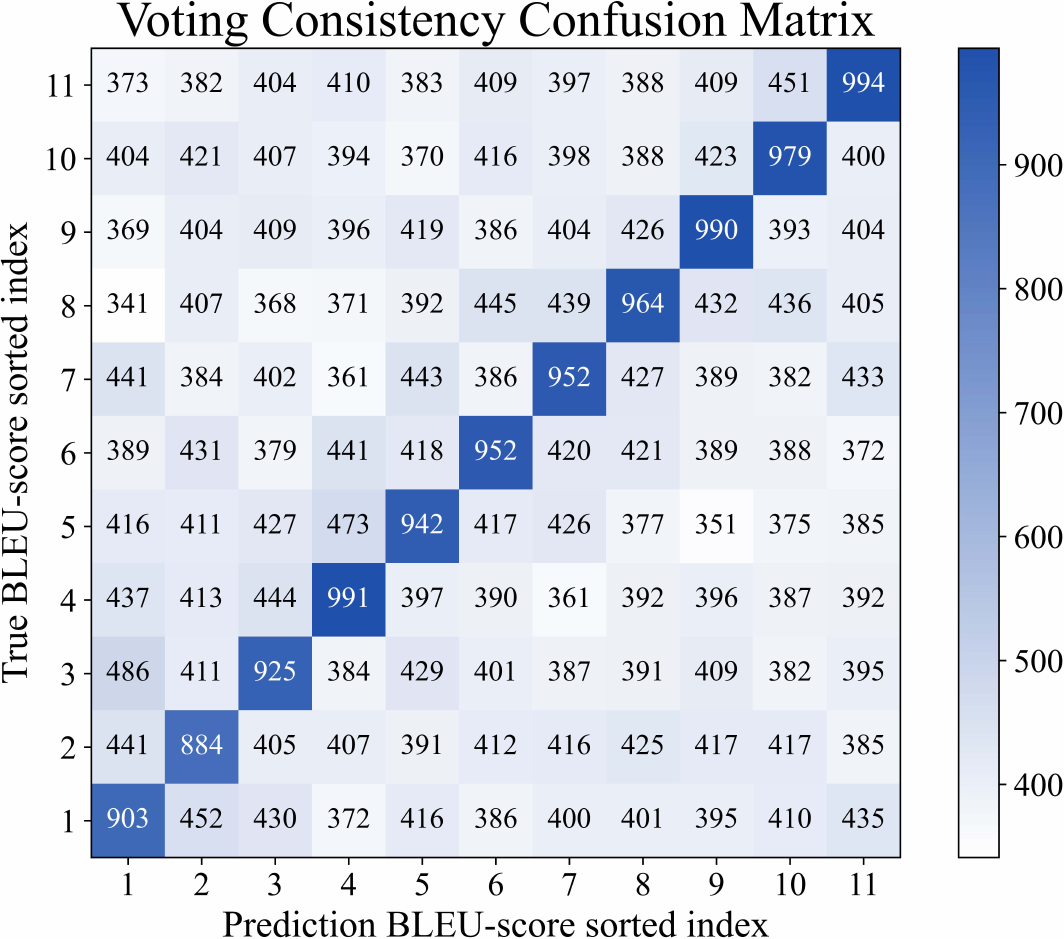}
    \caption{Confusion matrix for BLEU score.}
  \end{subfigure}
  \hfill
  \begin{subfigure}{0.32\textwidth}
    \includegraphics[width=\linewidth]{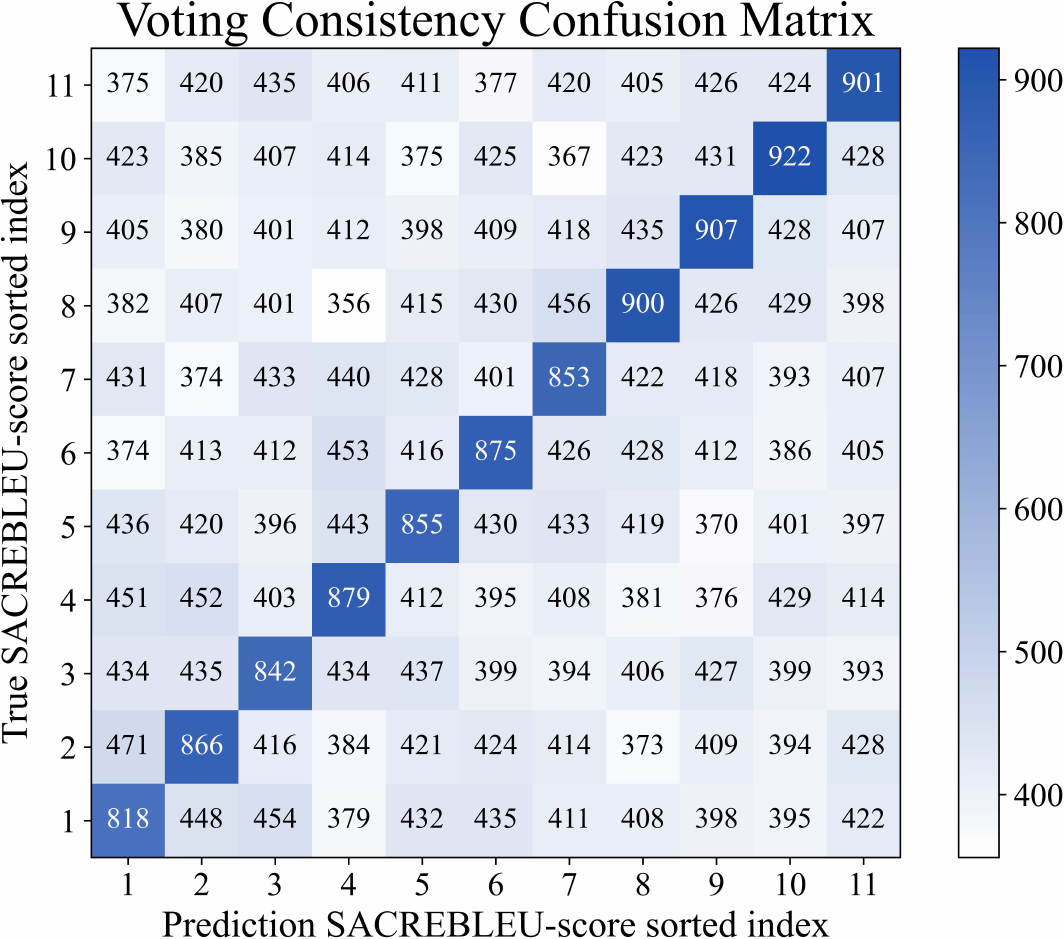}
    \caption{Confusion matrix for SacreBLEU score.}
  \end{subfigure}
  \hfill
  \begin{subfigure}{0.32\textwidth}
    \includegraphics[width=\linewidth]{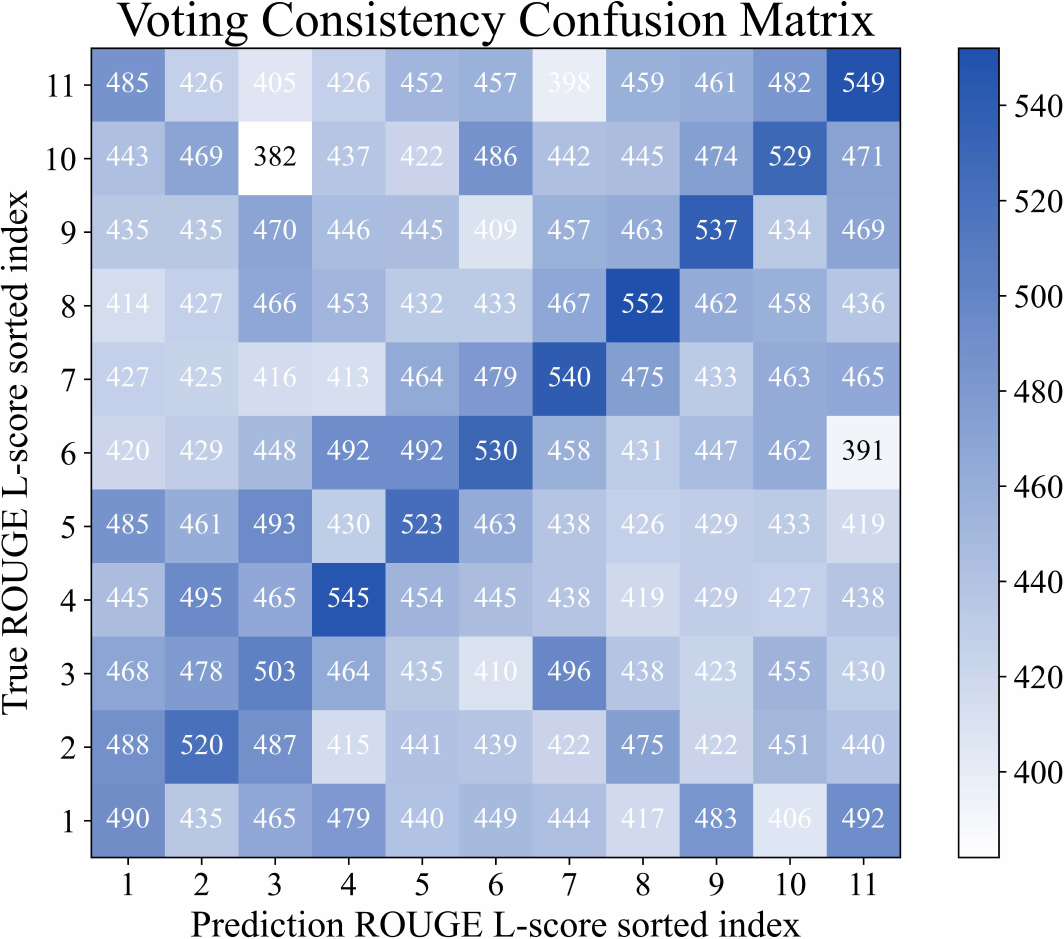}
    \caption{Confusion matrix for ROUGE L score.}
  \end{subfigure}

  \caption{Confusion matrix of Self-consistency Ensemble Voting for 5,000 test samples. The horizontal axis represents the index sorted from smallest to largest based on the scores calculated between each sample's top 11 retrieval-augmented prediction results and the ground truth, while the vertical axis represents the index sorted from smallest to largest based on the confidence levels obtained from voting among these 11 predictions for each sample, using cosine similarity for the voting process.}
  \label{confusion}
\vspace{-0.1in}
\end{figure*}
\subsubsection{Qualitative Results} 
\vspace*{-0.1in}
\hspace{3mm} Figure \ref{quality} presents the qualitative results comparison between our proposed model and other models including LLaVA-FT, inverse cooking \cite{Inverse_Cooking}, FoodLMM \cite{FoodLMM}, as well as the ground truth ``GT". It can be observed that our model, compared to the other models, can predict ingredients, preparation time, and temperature details more accurately, and produce more detailed and precise instructions. For instance, for Figure \ref{quality} (a), our model successfully identifies ingredients that the LLaVA-FT model fails to recognize, such as `milk', `butter' and `onions', and it makes more precise predictions regarding the timing, especially noting ``10 to 15 minutes." For Figure \ref{quality} (b), the generated recipe by our model is quite semantically similar to the GT, whereas the other models erroneously predicted unnecessary ingredients, such as `flour', `vanilla' and `nuts' in FoodLMM. These results demonstrate the effectiveness of our model to alleviate the issue of hallucination in recipe generation. 
Figure \ref{bad_case} shows that our model, while able to identify some correct ingredients like `tomatoes’ and `garlic’ for the food `tuna and tomato cream pasta sauce', still performs poorly in generating comprehensive instructions. A significant reason is our textual evaluation metrics cannot recognize `pasta' and `spaghetti’ as the same food; hence, even if the model correctly identifies `pasta' as ‘spaghetti’, the difference in expression leads to low textual metric scores. 
Similarly, leading models also struggle to predict accurate recipes. We aim to improve our model's ingredient recognition and text generation to match the training dataset's vocabulary and style for more robust application across various recipes.

\vspace{-0.1in}
\subsection{Ablation Study}
\begin{table}[]
\centering
\setlength{\abovecaptionskip}{1pt}
\caption{Comparison of ingredient recognition results in terms of IOU and F1.}
\label{recipe1m}
\begin{tabular}{c|c|c}
\hline
Methods & IOU(\%)         & F1(\%)        \\ \hline
$R_{I2L}$ \cite{salvador2017learning} & 18.92  &31.83 \\ \hline
$R_{I2LR}$ \cite{salvador2017learning} &19.85  & 33.13 \\ \hline
$FF_{TD}$ \cite{Inverse_Cooking} & 29.82  &45.94 \\ \hline
InverseCooking \cite{Inverse_Cooking} &32.11  &48.61 \\ \hline
FIRE \cite{FIRE}    & 32.59          & 49.27          \\ \hline
\textbf{Ours}    & \textbf{33.62} & \textbf{50.32} \\ \hline
\end{tabular}
\vspace{-0.1in}
\end{table}
\subsubsection{Stochastic Diversified Retrieval Augmentation (SDRA)} 
\vspace*{-0.1in}
\hspace*{1em}We first investigate the effect number of retrieved recipes $K$ for randomization for SDRA. 
Note that we do not use Self-consistency Ensemble Voting during inference to ensure the fairness of the experiment. Instead, SDRA is directly added to LLaVA-FT by increasing $K$ from 1, 10, 50 to 100 to investigate the effect
of the number of retrieved recipes for randomization. 
The results displayed in Table \ref{ablation_rag} indicate
that SDRA (top 50) performs the best, demonstrating that
our SDRA enhances its effectiveness by increasing the 
diversity and richness of the retrieved recipes within a certain
retrieval scope. However, expanding the scope of the search
too broadly can introduce noise into the model, thereby diminishing its performance. Specifically, for the $k=1$ case, the top 1 and top 2 retrieved sets of ingredients are added before $Q_{ingredients}$, while the top 3 ingredients and retrieved instructions are concatenated before $Q_{instructions}$. 
The results verify the effectiveness of our SDRA by increasing the diversity and richness of the retrieved recipes.
\begin{table}[]
\centering
\setlength{\abovecaptionskip}{1pt}
\caption{Ablation study of Stochastic Diversified Retrieval Augmentation (SDRA). "LLaVA-FT" denotes fine-tuned LLaVA, which is used as our baseline. "SDRA(fixed top 1)" refers to the approach where specifically, the top 2 retrieved ingredients sets and the top 3 ingredients along with top 1 instruction are pre-appended to their respective query placeholders. "SDRA(top k)" refers to the augmentation of the model by randomly selecting top k retrieval information when using the SDRA method.}
\label{ablation_rag}
\scriptsize
\begin{tabular}{cl|ccc}
\hline
\multicolumn{2}{c|}{Methods}          & BLEU  & SacreBLEU & ROUGE L \\ \hline
LLaVA-FT & \hspace{-1em}                                                                      & 28.32 & 5.88      & 38.18   \\ \hline
         & \hspace{-1em}+SDRA(fixed top 1) & 28.79 & 6.08      & \textbf{38.46}   \\ \hline
         & \hspace{-1em}+SDRA(top 10)      & 28.52 & 6.07      & \textbf{38.46}   \\ \hline
         & \hspace{-1em}+\textbf{SDRA(top 50)}      & \textbf{29.23} & \textbf{6.21}      & 38.43   \\ \hline
         & \hspace{-1em}+SDRA(top 100)      & 28.67 & 6.04      & 38.36   \\ \hline
\end{tabular}
\vspace{-0.1in}
\end{table}

\begin{figure}[h]
  \centering
  \includegraphics[width=\linewidth]{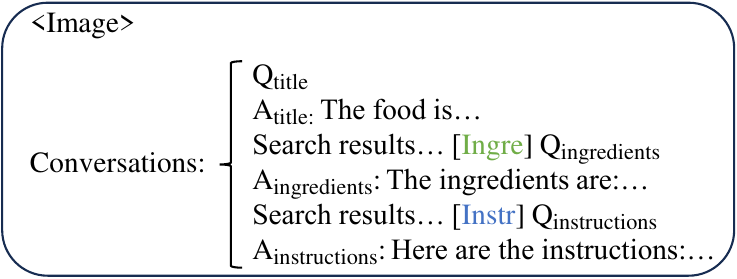}
  \caption{One set for Recipe demonstration $R$.} 
  \label{prompt_templet_1}
\vspace{-0.1in}
\end{figure}

Furthermore, we adopted two methods for concatenating retrieved information to verify the impact of the amount of retrieved information on the model's performance. The first method is as shown in Figure \ref{framework} (a), where 3 sets of ingredients and 1 set of instructions are added. The second method involves adding only 1 set of retrieved ingredients before $Q_{ingredients}$, and similarly, adding only 1 set of instructions before $Q_{instructions}$ as shown in Figure \ref{prompt_templet_1}. Figure \ref{prompt_templet_1} shows the format of the training data. Although the title is not required during the inference process, we include it during training to increase the amount of information used for training. During the inference phase, only 1 set of instructions is added before $Q_{instructions}$. In this way, we compare the use of the traditional RAG method, which involves adding fixed retrieval information before $Q_{ingredients}$ and $Q_{instructions}$. Specifically, the top 1 and top 2 retrieved sets of ingredients are added before $Q_{ingredients}$, while the top 3 ingredients and retrieved instructions are concatenated before $Q_{instructions}$. Table \ref{ablation_2set} indicates that incorporating 2 sets of retrieved recipes is more beneficial for the model's predictions than that of 1 set. However, due to the length limitations of model input, we are not able to examine more than two sets before each of the question.

\vspace{-0.1in}
\begin{table}[]
\centering
\captionsetup{skip=1pt}
\caption{Ablation study of the way concatenating retrieved information. 
``(1 set)" and ``(2 sets)" indicate Recipe demonstration $R$ as shown in Figure \ref{prompt_templet_1} and
 Figure \ref{framework} (a) respectively.}
\label{ablation_2set}
\scriptsize
\begin{tabular}{cl|ccc}
\hline
\multicolumn{2}{c|}{Methods}          & BLEU  & SacreBLEU & ROUGE L \\ \hline
LLaVA-FT & \hspace{-1em}                                                                      & 28.32 & 5.88      & 38.18   \\ \hline
         & \hspace{-1em}+SDRA(1 set) & 28.79 & 6.08      & \textbf{38.46}   \\ \hline
         & \hspace{-1em}+\textbf{SDRA(2 sets)}      & \textbf{29.23} & \textbf{6.21}      & 38.43   \\ \hline
\end{tabular}
\vspace{-0.1in}
\end{table}


\begin{table}[]
\centering
\setlength{\abovecaptionskip}{1pt}
\caption{Ablation study of Self-consistency Ensemble Voting. 
`S' refers to the number of generated recipes for ensemble voting. `Sum' is the sum of cosine similarity scores.}
\label{ablation_voting}
\scriptsize
\begin{tabular}{c|c|cccc}
\hline
Scoring metric                                                                & Number & BLEU           & SacreBLEU     & ROUGE L        & Sum            \\ \hline
\multirow{6}{*}{\begin{tabular}[c]{@{}c@{}}Cosine \\ Similarity\end{tabular}} & S=1    & 29.23          & 6.21          & 38.43          & 73.87          \\ \cline{2-6} 
                                                                              & S=3    & 29.68          & 6.31          & 38.68          & 74.67          \\ \cline{2-6} 
                                                                              & S=5    & 30.12          & 6.39          & 38.66          & 75.17          \\ \cline{2-6} 
                                                                              & S=7    & 30.07          & 6.41          & 38.84          & 75.32          \\ \cline{2-6} 
                                                                              & S=9    & \textbf{30.11} & \textbf{6.42} & 38.91          & 75.44          \\ \cline{2-6} 
                                                                              & S=11   & \textbf{30.11} & \textbf{6.42} & \textbf{38.93} & \textbf{75.47} \\ \hline
\end{tabular}
\vspace{-0.1in}
\end{table}

\subsubsection{Self-consistency Ensemble Voting}
\vspace*{-0.1in}
\hspace*{1em}Table \ref{ablation_voting} shows the variation in recipe generation quality when calculating the confidence of candidate recipes using cosine similarity. The table shows that as more candidate recipes are generated, the text metrics for recipe quality steadily improve. In addition to cosine similarity, we also used BLEU, SacreBLEU, and ROUGE L as scoring metrics, with results presented in Section 4 of the supplementary materials. Our best result, as displayed in Table \ref{sota}, is based on Cosine Similarity due to its consistent performance. We choose S=11 because the BLEU and SacreBLEU stopped improving at S=11. These results verify the effectiveness of our Self-consistency Ensemble Voting. As cosine similarity shows more steady and robust results in improving BLEU, SacreBLEU, and ROUGE L scores with the increase of $S$, we report the results of cosine similarity with $S=11$ in Table~\ref{sota}. As the parameter S increases, the computational cost scales linearly. We consistently observe performance improvements as S grows. However, this comes with a trade-off between performance gains and computational expense. For instance, while higher values like S = 11 may yield better results, selecting S = 5 could be a more practical choice when computational resources are constrained, as it strikes a balance between efficiency and performance.

Additionally, in Figure \ref{confusion}, we plot a confusion matrix comparing the confidence ranking of seven predictions—obtained through voting with cosine similarity on the top 11 retrieval results from 5,000 test samples—with their actual BLEU, SacreBLEU, and ROUGE L score rankings. The shade of color represents the number of samples, for example, in Figure \ref{confusion} (a), (1,3) indicates the number of samples with the highest confidence yet ranked third in BLEU scores. The results show that the final prediction selected by confidence and the actual prediction scores are consistent, leading to consistently better performance, further emphasizing the importance of introducing the voting mechanism during inference. 
\vspace{-0.1in}
\section{Conclusion}
We have presented the first retrieval augmented large multimodal mode to mitigate the hallucination issue for recipe generation. We introduce the Stochastic Diversified Retrieval Augmentation to enable the model to better acquire useful knowledge from retrieved retrieval and propose Self-consistency Ensemble Voting, which optimizes the final instructions by scoring predictions obtained from different retrieval information against each other. 
Experimental results validate our method's effectiveness and potential for widespread application in food computing. Future work will introduce a self-reflection strategy to refine incorrect generation results, improving recipe accuracy and reliability.
\section*{Acknowledgements}
\noindent This research/project is supported by the Ministry of Education, Singapore, under its Academic Research Fund Tier 2 (Proposal ID: T2EP20222-0046). Any opinions, findings and conclusions or recommendations expressed in this material
are those of the authors and do not reflect the views of the Ministry of Education,
Singapore.
{\small
\bibliographystyle{ieee_fullname}
\bibliography{egbib}

\begin{thebibliography}{10}\itemsep=-1pt

\bibitem{gpt4}
Josh Achiam, Steven Adler, Sandhini Agarwal, Lama Ahmad, Ilge Akkaya, Florencia~Leoni Aleman, Diogo Almeida, Janko Altenschmidt, Sam Altman, Shyamal Anadkat, et~al.
\newblock Gpt-4 technical report.
\newblock {\em arXiv preprint arXiv:2303.08774}, 2023.

\bibitem{Flamingo}
Jean-Baptiste Alayrac, Jeff Donahue, Pauline Luc, Antoine Miech, Iain Barr, Yana Hasson, Karel Lenc, Arthur Mensch, Katherine Millican, Malcolm Reynolds, et~al.
\newblock Flamingo: a visual language model for few-shot learning.
\newblock {\em Advances in neural information processing systems}, 35:23716--23736, 2022.

\bibitem{Food-101}
Lukas Bossard, Matthieu Guillaumin, and Luc Van~Gool.
\newblock Food-101--mining discriminative components with random forests.
\newblock In {\em Computer Vision--ECCV 2014: 13th European Conference, Zurich, Switzerland, September 6-12, 2014, Proceedings, Part VI 13}, pages 446--461. Springer, 2014.

\bibitem{VisualGPT}
Jun Chen, Han Guo, Kai Yi, Boyang Li, and Mohamed Elhoseiny.
\newblock Visualgpt: Data-efficient adaptation of pretrained language models for image captioning.
\newblock In {\em Proceedings of the IEEE/CVF Conference on Computer Vision and Pattern Recognition}, pages 18030--18040, 2022.

\bibitem{chen2024benchmarking}
Jiawei Chen, Hongyu Lin, Xianpei Han, and Le Sun.
\newblock Benchmarking large language models in retrieval-augmented generation.
\newblock In {\em Proceedings of the AAAI Conference on Artificial Intelligence}, volume~38, pages 17754--17762, 2024.

\bibitem{Deep-based}
Jingjing Chen and Chong-Wah Ngo.
\newblock Deep-based ingredient recognition for cooking recipe retrieval.
\newblock In {\em Proceedings of the 24th ACM international conference on Multimedia}, pages 32--41, 2016.

\bibitem{fvl1}
Jingjing Chen, Bin Zhu, Chong-Wah Ngo, Tat-Seng Chua, and Yu-Gang Jiang.
\newblock A study of multi-task and region-wise deep learning for food ingredient recognition.
\newblock {\em IEEE Transactions on Image Processing}, 30:1514--1526, 2020.

\bibitem{MiniGPT-v2}
Jun Chen, Deyao Zhu, Xiaoqian Shen, Xiang Li, Zechun Liu, Pengchuan Zhang, Raghuraman Krishnamoorthi, Vikas Chandra, Yunyang Xiong, and Mohamed Elhoseiny.
\newblock Minigpt-v2: large language model as a unified interface for vision-language multi-task learning.
\newblock {\em arXiv preprint arXiv:2310.09478}, 2023.

\bibitem{Cross-modal}
Jing-jing Chen, Chong-Wah Ngo, and Tat-Seng Chua.
\newblock Cross-modal recipe retrieval with rich food attributes.
\newblock In {\em Proceedings of the 25th ACM international conference on Multimedia}, pages 1771--1779, 2017.

\bibitem{VAL}
Yanbei Chen, Shaogang Gong, and Loris Bazzani.
\newblock Image search with text feedback by visiolinguistic attention learning.
\newblock In {\em Proceedings of the IEEE/CVF Conference on Computer Vision and Pattern Recognition}, pages 3001--3011, 2020.

\bibitem{FIRE}
Prateek Chhikara, Dhiraj Chaurasia, Yifan Jiang, Omkar Masur, and Filip Ilievski.
\newblock Fire: Food image to recipe generation.
\newblock pages 8184--8194, 2024.

\bibitem{ciocca2016food}
Gianluigi Ciocca, Paolo Napoletano, and Raimondo Schettini.
\newblock Food recognition: a new dataset, experiments, and results.
\newblock {\em IEEE journal of biomedical and health informatics}, 21(3):588--598, 2016.

\bibitem{cox2017food}
Andrew~Martin Cox, Pamela McKinney, and Paula Goodale.
\newblock Food logging: an information literacy perspective.
\newblock {\em Aslib Journal of Information Management}, 69(2):184--200, 2017.

\bibitem{Instructblip}
Wenliang Dai, Junnan Li, Dongxu Li, Anthony Meng~Huat Tiong, Junqi Zhao, Weisheng Wang, Boyang Li, Pascale~N Fung, and Steven Hoi.
\newblock Instructblip: Towards general-purpose vision-language models with instruction tuning.
\newblock {\em Advances in Neural Information Processing Systems}, 36, 2024.

\bibitem{BERT}
Jacob Devlin, Ming-Wei Chang, Kenton Lee, and Kristina Toutanova.
\newblock Bert: Pre-training of deep bidirectional transformers for language understanding.
\newblock 2018.

\bibitem{VIT}
Alexey Dosovitskiy, Lucas Beyer, Alexander Kolesnikov, Dirk Weissenborn, Xiaohua Zhai, Thomas Unterthiner, Mostafa Dehghani, Matthias Minderer, Georg Heigold, Sylvain Gelly, et~al.
\newblock An image is worth 16x16 words: Transformers for image recognition at scale.
\newblock 2020.

\bibitem{elsweiler2017exploiting}
David Elsweiler, Christoph Trattner, and Morgan Harvey.
\newblock Exploiting food choice biases for healthier recipe recommendation.
\newblock In {\em Proceedings of the 40th international acm sigir conference on research and development in information retrieval}, pages 575--584, 2017.

\bibitem{T5}
Mehrdad Farahani, Kartik Godawat, Haswanth Aekula, Deepak Pandian, and Nicholas Broad.
\newblock Chef transformer, 2023.

\bibitem{freyne2010intelligent}
Jill Freyne and Shlomo Berkovsky.
\newblock Intelligent food planning: personalized recipe recommendation.
\newblock In {\em Proceedings of the 15th international conference on Intelligent user interfaces}, pages 321--324, 2010.

\bibitem{FIGURE}
Saadia Gabriel, Asli Celikyilmaz, Rahul Jha, Yejin Choi, and Jianfeng Gao.
\newblock Go figure: A meta evaluation of factuality in summarization.
\newblock 2020.

\bibitem{gao2023retrieval}
Yunfan Gao, Yun Xiong, Xinyu Gao, Kangxiang Jia, Jinliu Pan, Yuxi Bi, Yi Dai, Jiawei Sun, and Haofen Wang.
\newblock Retrieval-augmented generation for large language models: A survey.
\newblock {\em arXiv preprint arXiv:2312.10997}, 2023.

\bibitem{gyx}
Yinxuan Gui, Bin Zhu, Jingjing Chen, Chong~Wah Ngo, and Yu-Gang Jiang.
\newblock Navigating weight prediction with diet diary.
\newblock In {\em Proceedings of the 32nd ACM International Conference on Multimedia}, page 127–136, 2024.

\bibitem{Multi-Modal_Classification}
Shir Gur, Natalia Neverova, Chris Stauffer, Ser-Nam Lim, Douwe Kiela, and Austin Reiter.
\newblock Cross-modal retrieval augmentation for multi-modal classification.
\newblock 2021.

\bibitem{REALM}
Kelvin Guu, Kenton Lee, Zora Tung, Panupong Pasupat, and Mingwei Chang.
\newblock Retrieval augmented language model pre-training.
\newblock pages 3929--3938, 2020.

\bibitem{Recipegpt}
Helena H.~Lee, Ke Shu, Palakorn Achananuparp, Philips~Kokoh Prasetyo, Yue Liu, Ee-Peng Lim, and Lav~R Varshney.
\newblock Recipegpt: Generative pre-training based cooking recipe generation and evaluation system.
\newblock In {\em Companion Proceedings of the Web Conference 2020}, pages 181--184, 2020.

\bibitem{lora}
Edward~J Hu, Yelong Shen, Phillip Wallis, Zeyuan Allen-Zhu, Yuanzhi Li, Shean Wang, Lu Wang, and Weizhu Chen.
\newblock Lora: Low-rank adaptation of large language models.
\newblock 2021.

\bibitem{REVEAL}
Ziniu Hu, Ahmet Iscen, Chen Sun, Zirui Wang, Kai-Wei Chang, Yizhou Sun, Cordelia Schmid, David~A Ross, and Alireza Fathi.
\newblock Reveal: Retrieval-augmented visual-language pre-training with multi-source multimodal knowledge memory.
\newblock In {\em Proceedings of the IEEE/CVF conference on computer vision and pattern recognition}, pages 23369--23379, 2023.

\bibitem{vireosota}
Shuqiang Jiang, Weiqing Min, Linhu Liu, and Zhengdong Luo.
\newblock Multi-scale multi-view deep feature aggregation for food recognition.
\newblock {\em IEEE Transactions on Image Processing}, 29:265--276, 2019.

\bibitem{Active_Retrieval_Augmented_Generation}
Zhengbao Jiang, Frank~F Xu, Luyu Gao, Zhiqing Sun, Qian Liu, Jane Dwivedi-Yu, Yiming Yang, Jamie Callan, and Graham Neubig.
\newblock Active retrieval augmented generation.
\newblock 2023.

\bibitem{jpk}
Pengkun Jiao, Xinlan Wu, Bin Zhu, Jingjing Chen, Chong-Wah Ngo, and Yugang Jiang.
\newblock Rode: Linear rectified mixture of diverse experts for food large multi-modal models.
\newblock {\em arXiv preprint arXiv:2407.12730}, 2024.

\bibitem{jiao2024lumen}
Yang Jiao, Shaoxiang Chen, Zequn Jie, Jingjing Chen, Lin Ma, and Yu-Gang Jiang.
\newblock Lumen: Unleashing versatile vision-centric capabilities of large multimodal models.
\newblock {\em arXiv preprint arXiv:2403.07304}, 2024.

\bibitem{jiao2021two}
Yang Jiao, Zequn Jie, Weixin Luo, Jingjing Chen, Yu-Gang Jiang, Xiaolin Wei, and Lin Ma.
\newblock Two-stage visual cues enhancement network for referring image segmentation.
\newblock In {\em Proceedings of the 29th ACM international conference on multimedia}, pages 1331--1340, 2021.

\bibitem{lisa}
Xin Lai, Zhuotao Tian, Yukang Chen, Yanwei Li, Yuhui Yuan, Shu Liu, and Jiaya Jia.
\newblock Lisa: Reasoning segmentation via large language model.
\newblock {\em arXiv preprint arXiv:2308.00692}, 2023.

\bibitem{DBLP:conf/nips/LewisGGAWZ20}
Mike Lewis, Marjan Ghazvininejad, Gargi Ghosh, Armen Aghajanyan, Sida Wang, and Luke Zettlemoyer.
\newblock Pre-training via paraphrasing.
\newblock volume~33, pages 18470--18481, 2020.

\bibitem{DBLP:conf/nips/LewisPPPKGKLYR020}
Patrick Lewis, Ethan Perez, Aleksandra Piktus, Fabio Petroni, Vladimir Karpukhin, Naman Goyal, Heinrich K{\"u}ttler, Mike Lewis, Wen-tau Yih, Tim Rockt{\"a}schel, et~al.
\newblock Retrieval-augmented generation for knowledge-intensive nlp tasks.
\newblock volume~33, pages 9459--9474, 2020.

\bibitem{Knowledge-Intensive}
Patrick Lewis, Ethan Perez, Aleksandra Piktus, Fabio Petroni, Vladimir Karpukhin, Naman Goyal, Heinrich K{\"u}ttler, Mike Lewis, Wen-tau Yih, Tim Rockt{\"a}schel, et~al.
\newblock Retrieval-augmented generation for knowledge-intensive nlp tasks.
\newblock volume~33, pages 9459--9474, 2020.

\bibitem{Llava-med}
Chunyuan Li, Cliff Wong, Sheng Zhang, Naoto Usuyama, Haotian Liu, Jianwei Yang, Tristan Naumann, Hoifung Poon, and Jianfeng Gao.
\newblock Llava-med: Training a large language-and-vision assistant for biomedicine in one day.
\newblock {\em Advances in Neural Information Processing Systems}, 36, 2024.

\bibitem{BLIP-2}
Junnan Li, Dongxu Li, Silvio Savarese, and Steven Hoi.
\newblock Blip-2: Bootstrapping language-image pre-training with frozen image encoders and large language models.
\newblock In {\em International conference on machine learning}, pages 19730--19742. PMLR, 2023.

\bibitem{BLIP}
Junnan Li, Dongxu Li, Caiming Xiong, and Steven Hoi.
\newblock Blip: Bootstrapping language-image pre-training for unified vision-language understanding and generation.
\newblock In {\em International conference on machine learning}, pages 12888--12900. PMLR, 2022.

\bibitem{Evaluating}
Yifan Li, Yifan Du, Kun Zhou, Jinpeng Wang, Wayne~Xin Zhao, and Ji-Rong Wen.
\newblock Evaluating object hallucination in large vision-language models.
\newblock 2023.

\bibitem{li2024eyes}
Yian Li, Wentao Tian, Yang Jiao, and Jingjing Chen.
\newblock Eyes can deceive: Benchmarking counterfactual reasoning abilities of multi-modal large language models.
\newblock {\em arXiv preprint arXiv:2404.12966}, 2024.

\bibitem{liu2020food}
Chengxu Liu, Yuanzhi Liang, Yao Xue, Xueming Qian, and Jianlong Fu.
\newblock Food and ingredient joint learning for fine-grained recognition.
\newblock {\em IEEE transactions on circuits and Systems for Video Technology}, 31(6):2480--2493, 2020.

\bibitem{lgs}
Guoshan Liu, Yang Jiao, Jingjing Chen, Bin Zhu, and Yu-Gang Jiang.
\newblock From canteen food to daily meals: Generalizing food recognition to more practical scenarios.
\newblock {\em IEEE Transactions on Multimedia}, pages 1--10, 2024.

\bibitem{LLaVA}
Haotian Liu, Chunyuan Li, Qingyang Wu, and Yong~Jae Lee.
\newblock Visual instruction tuning.
\newblock volume~36, 2024.

\bibitem{liu2022transformer}
Xinda Liu, Lili Wang, and Xiaoguang Han.
\newblock Transformer with peak suppression and knowledge guidance for fine-grained image recognition.
\newblock {\em Neurocomputing}, 492:137--149, 2022.

\bibitem{majumder2019generating}
Bodhisattwa~Prasad Majumder, Shuyang Li, Jianmo Ni, and Julian McAuley.
\newblock Generating personalized recipes from historical user preferences.
\newblock {\em arXiv preprint arXiv:1909.00105}, 2019.

\bibitem{DBLP:conf/wacv/MartinelFM18}
Niki Martinel, Gian~Luca Foresti, and Christian Micheloni.
\newblock Wide-slice residual networks for food recognition.
\newblock In {\em 2018 IEEE Winter conference on applications of computer vision (WACV)}, pages 567--576. IEEE, 2018.

\bibitem{food101sota}
Sachit Menon and Carl Vondrick.
\newblock Visual classification via description from large language models.
\newblock 2022.

\bibitem{food_survey}
Weiqing Min, Shuqiang Jiang, Linhu Liu, Yong Rui, and Ramesh Jain.
\newblock A survey on food computing.
\newblock {\em ACM Computing Surveys (CSUR)}, 52(5):1--36, 2019.

\bibitem{Food2K}
Weiqing Min, Zhiling Wang, Yuxin Liu, Mengjiang Luo, Liping Kang, Xiaoming Wei, Xiaolin Wei, and Shuqiang Jiang.
\newblock Large scale visual food recognition.
\newblock {\em IEEE Transactions on Pattern Analysis and Machine Intelligence}, 2023.

\bibitem{Hierarchical}
Hai~X Pham, Ricardo Guerrero, Vladimir Pavlovic, and Jiatong Li.
\newblock Chef: cross-modal hierarchical embeddings for food domain retrieval.
\newblock In {\em Proceedings of the AAAI Conference on Artificial Intelligence}, volume~35, pages 2423--2430, 2021.

\bibitem{In-Context}
Ori Ram, Yoav Levine, Itay Dalmedigos, Dor Muhlgay, Amnon Shashua, Kevin Leyton-Brown, and Yoav Shoham.
\newblock In-context retrieval-augmented language models.
\newblock {\em Transactions of the Association for Computational Linguistics}, 11:1316--1331, 2023.

\bibitem{smallcap}
Rita Ramos, Bruno Martins, Desmond Elliott, and Yova Kementchedjhieva.
\newblock Smallcap: lightweight image captioning prompted with retrieval augmentation.
\newblock In {\em Proceedings of the IEEE/CVF Conference on Computer Vision and Pattern Recognition}, pages 2840--2849, 2023.

\bibitem{rodenas2022learning}
Javier R{\'o}denas, Bhalaji Nagarajan, Marc Bola{\~n}os, and Petia Radeva.
\newblock Learning multi-subset of classes for fine-grained food recognition.
\newblock In {\em Proceedings of the 7th International Workshop on Multimedia Assisted Dietary Management}, pages 17--26, 2022.

\bibitem{sahoo2019foodai}
Doyen Sahoo, Wang Hao, Shu Ke, Wu Xiongwei, Hung Le, Palakorn Achananuparp, Ee-Peng Lim, and Steven~CH Hoi.
\newblock Foodai: Food image recognition via deep learning for smart food logging.
\newblock In {\em Proceedings of the 25th ACM SIGKDD International Conference on Knowledge Discovery \& Data Mining}, pages 2260--2268, 2019.

\bibitem{Inverse_Cooking}
Amaia Salvador, Michal Drozdzal, Xavier Gir{\'o}-i Nieto, and Adriana Romero.
\newblock Inverse cooking: Recipe generation from food images.
\newblock In {\em Proceedings of the IEEE/CVF Conference on Computer Vision and Pattern Recognition}, pages 10453--10462, 2019.

\bibitem{Revamping}
Amaia Salvador, Erhan Gundogdu, Loris Bazzani, and Michael Donoser.
\newblock Revamping cross-modal recipe retrieval with hierarchical transformers and self-supervised learning.
\newblock In {\em Proceedings of the IEEE/CVF Conference on Computer Vision and Pattern Recognition}, pages 15475--15484, 2021.

\bibitem{salvador2017learning}
Amaia Salvador, Nicholas Hynes, Yusuf Aytar, Javier Marin, Ferda Ofli, Ingmar Weber, and Antonio Torralba.
\newblock Learning cross-modal embeddings for cooking recipes and food images.
\newblock In {\em Proceedings of the IEEE conference on computer vision and pattern recognition}, pages 3020--3028, 2017.

\bibitem{sfz}
Fangzhou Song, Bin Zhu, Yanbin Hao, and Shuo Wang.
\newblock Enhancing recipe retrieval with foundation models: A data augmentation perspective.
\newblock In {\em European Conference on Computer Vision}, pages 111--127, 2024.

\bibitem{Recipe_recommendation}
Chun-Yuen Teng, Yu-Ru Lin, and Lada~A Adamic.
\newblock Recipe recommendation using ingredient networks.
\newblock In {\em Proceedings of the 4th annual ACM web science conference}, pages 298--307, 2012.

\bibitem{LLAMA}
Hugo Touvron, Thibaut Lavril, Gautier Izacard, Xavier Martinet, Marie-Anne Lachaux, Timoth{\'e}e Lacroix, Baptiste Rozi{\`e}re, Naman Goyal, Eric Hambro, Faisal Azhar, et~al.
\newblock Llama: Open and efficient foundation language models.
\newblock {\em arXiv preprint arXiv:2302.13971}, 2023.

\bibitem{X-TRA}
Tom van Sonsbeek and Marcel Worring.
\newblock X-tra: Improving chest x-ray tasks with cross-modal retrieval augmentation.
\newblock In {\em International Conference on Information Processing in Medical Imaging}, pages 471--482. Springer, 2023.

\bibitem{TIRG}
Nam Vo, Lu Jiang, Chen Sun, Kevin Murphy, Li-Jia Li, Li Fei-Fei, and James Hays.
\newblock Composing text and image for image retrieval-an empirical odyssey.
\newblock In {\em Proceedings of the IEEE/CVF conference on computer vision and pattern recognition}, pages 6439--6448, 2019.

\bibitem{Structure-Aware}
Hao Wang, Guosheng Lin, Steven~CH Hoi, and Chunyan Miao.
\newblock Structure-aware generation network for recipe generation from images.
\newblock In {\em Computer Vision--ECCV 2020: 16th European Conference, Glasgow, UK, August 23--28, 2020, Proceedings, Part XXVII 16}, pages 359--374. Springer, 2020.

\bibitem{Structural_Representations}
Hao Wang, Guosheng Lin, Steven~CH Hoi, and Chunyan Miao.
\newblock Learning structural representations for recipe generation and food retrieval.
\newblock {\em IEEE Transactions on Pattern Analysis and Machine Intelligence}, 45(3):3363--3377, 2022.

\bibitem{recipe1m}
Hao Wang, Doyen Sahoo, Chenghao Liu, Ee-peng Lim, and Steven~CH Hoi.
\newblock Learning cross-modal embeddings with adversarial networks for cooking recipes and food images.
\newblock In {\em Proceedings of the IEEE/CVF conference on computer vision and pattern recognition}, pages 11572--11581, 2019.

\bibitem{GPT}
Qiaolin Xia, Haoyang Huang, Nan Duan, Dongdong Zhang, Lei Ji, Zhifang Sui, Edward Cui, Taroon Bharti, and Ming Zhou.
\newblock Xgpt: Cross-modal generative pre-training for image captioning.
\newblock In {\em Natural Language Processing and Chinese Computing: 10th CCF International Conference, NLPCC 2021, Qingdao, China, October 13--17, 2021, Proceedings, Part I 10}, pages 786--797. Springer, 2021.

\bibitem{yang2010food}
Shulin Yang, Mei Chen, Dean Pomerleau, and Rahul Sukthankar.
\newblock Food recognition using statistics of pairwise local features.
\newblock In {\em 2010 IEEE computer society conference on computer vision and pattern recognition}, pages 2249--2256. IEEE, 2010.

\bibitem{Retrieval-Augmented_Multimodal}
Michihiro Yasunaga, Armen Aghajanyan, Weijia Shi, Rich James, Jure Leskovec, Percy Liang, Mike Lewis, Luke Zettlemoyer, and Wen-tau Yih.
\newblock Retrieval-augmented multimodal language modeling.
\newblock 2022.

\bibitem{FoodLMM}
Yuehao Yin, Huiyan Qi, Bin Zhu, Jingjing Chen, Yu-Gang Jiang, and Chong-Wah Ngo.
\newblock Foodlmm: A versatile food assistant using large multi-modal model.
\newblock {\em arXiv preprint arXiv:2312.14991}, 2023.

\bibitem{zhang2024eagle}
Jiacheng Zhang, Yang Jiao, Shaoxiang Chen, Jingjing Chen, and Yu-Gang Jiang.
\newblock Eagle: Towards efficient arbitrary referring visual prompts comprehension for multimodal large language models.
\newblock {\em arXiv preprint arXiv:2409.16723}, 2024.

\bibitem{zhang2024eventhallusion}
Jiacheng Zhang, Yang Jiao, Shaoxiang Chen, Jingjing Chen, and Yu-Gang Jiang.
\newblock Eventhallusion: Diagnosing event hallucinations in video llms.
\newblock {\em arXiv preprint arXiv:2409.16597}, 2024.

\bibitem{zhang2022sequential}
Mengyang Zhang, Guohui Tian, Ying Zhang, and Hong Liu.
\newblock Sequential learning for ingredient recognition from images.
\newblock {\em IEEE Transactions on Circuits and Systems for Video Technology}, 2022.

\bibitem{vicuna}
Lianmin Zheng, Wei-Lin Chiang, Ying Sheng, Siyuan Zhuang, Zhanghao Wu, Yonghao Zhuang, Zi Lin, Zhuohan Li, Dacheng Li, Eric Xing, et~al.
\newblock Judging llm-as-a-judge with mt-bench and chatbot arena.
\newblock volume~36, 2024.

\bibitem{zhubin}
Bin Zhu, Chong-Wah Ngo, and Wing-Kwong Chan.
\newblock Learning from web recipe-image pairs for food recognition: Problem, baselines and performance.
\newblock {\em IEEE Transactions on Multimedia}, 24:1175--1185, 2021.

\bibitem{fvl4}
Bin Zhu, Chong-Wah Ngo, Jingjing Chen, and Yanbin Hao.
\newblock R2gan: Cross-modal recipe retrieval with generative adversarial network.
\newblock In {\em Proceedings of the IEEE/CVF Conference on Computer Vision and Pattern Recognition}, pages 11477--11486, 2019.

\bibitem{fvl2}
Bin Zhu, Chong-Wah Ngo, and Jing-jing Chen.
\newblock Cross-domain cross-modal food transfer.
\newblock In {\em Proceedings of the 28th ACM International Conference on Multimedia}, pages 3762--3770, 2020.

\end{thebibliography}
}

\end{document}


\title{Supplementary Material\\Retrieval Augmented Recipe Generation}

\author{
Guoshan Liu\textsuperscript{1,2}\thanks{Equal contribution.}, Hailong Yin\textsuperscript{1,2}$^*$, 
Bin Zhu\textsuperscript{3}, 
Jingjing Chen\textsuperscript{1,2}\thanks{Jingjing Chen is the corresponding author.}, 
Chong-Wah Ngo\textsuperscript{3}, 
Yu-Gang Jiang\textsuperscript{1,2} \\
\textsuperscript{1}Shanghai Key Lab of Intelligent Information Processing, School of Computer Science, Fudan University \\
\textsuperscript{2}Shanghai Collaborative Innovation Center on Intelligent Visual Computing \\
\textsuperscript{3}Singapore Management University \\
{\tt\small \{gsliu24, hlyin23\}@m.fudan.edu.cn, \{chenjingjing, ygj\}@fudan.edu.cn} \\
{\tt\small \{binzhu, cwngo\}@smu.edu.sg} 
}
\maketitle

\section{Details of Training Data Organization}

\begin{figure*}
  \centering
  \includegraphics[width=\linewidth]{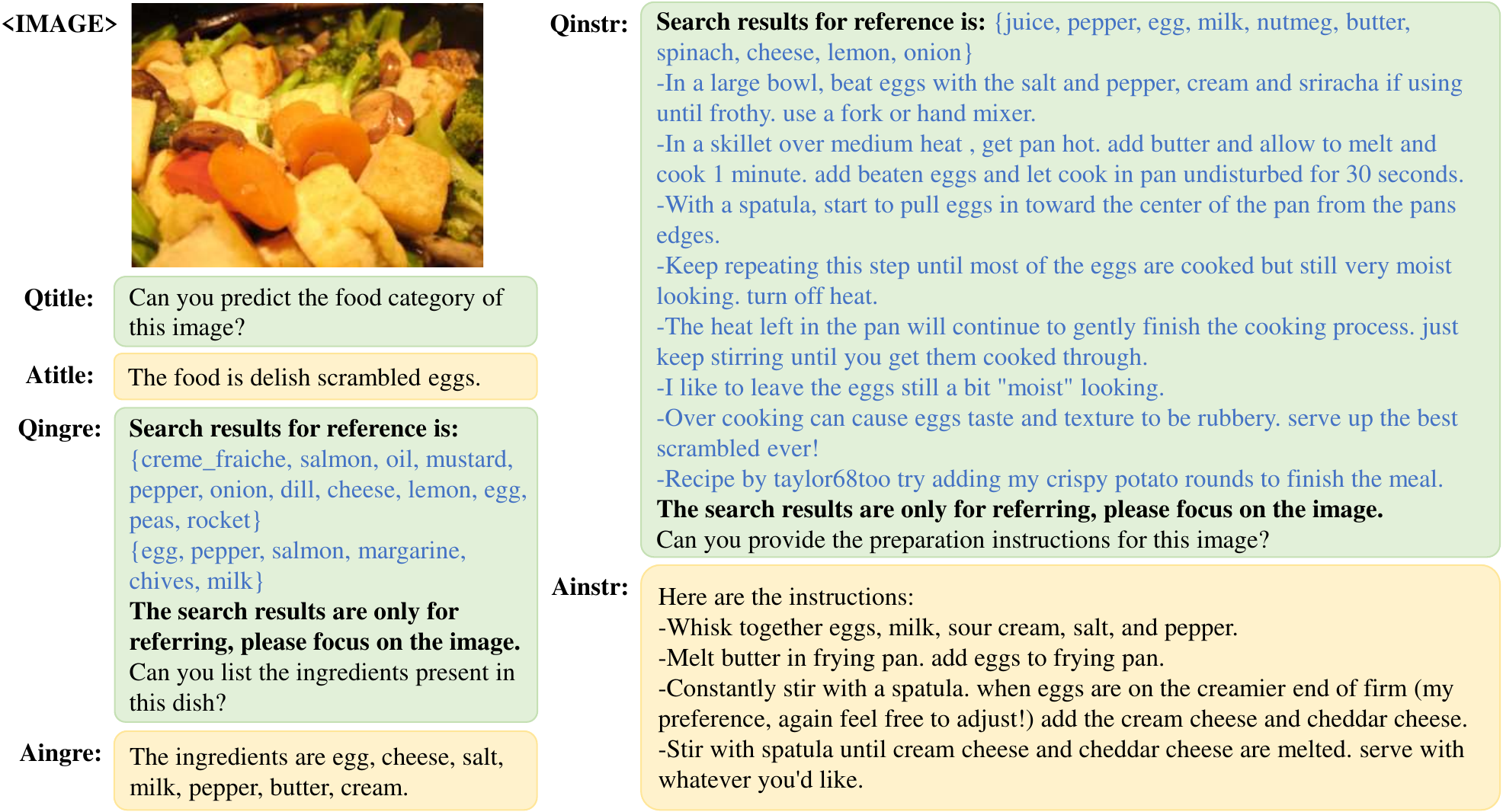}
  \caption{Our training data organization. \textbf{BOLD} serves as a fixed prompt, the blue font highlights the three groups of ingredients and one group of instructions retrieved, and the rest of the content represents the ground truth.}
  \label{training_data}
\end{figure*}

As illustrated in Figure \ref{training_data}, we present an example of how the training data is organized in our proposed retrieval augmented framework.

\section{Retrieved Results}
\begin{figure*}
  \centering
  \includegraphics[width=\linewidth]{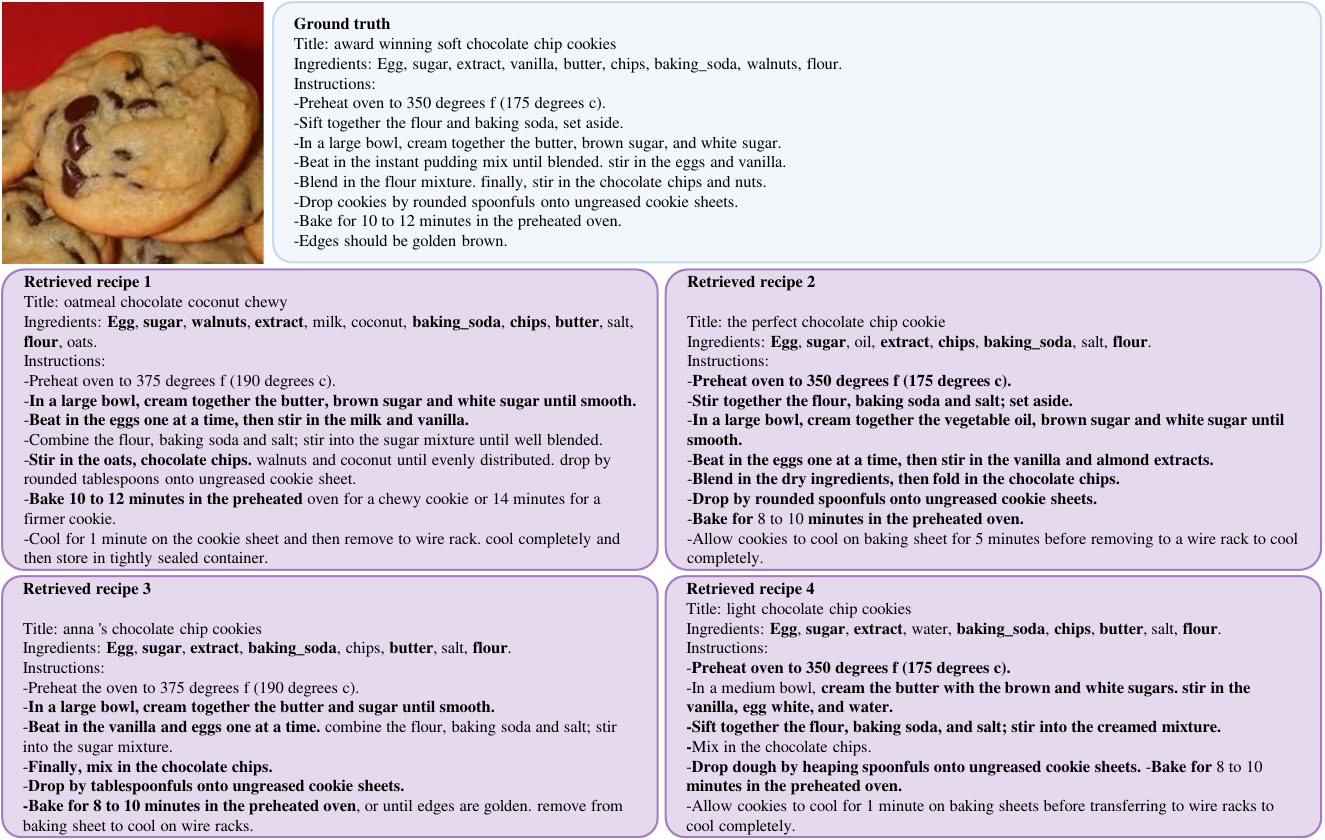}
  \captionsetup{skip=1pt}
  \caption{Comparison between retrieval results and ground truth. BOLD represents the relevant and similar parts between the retrieved results and the ground truth. This is the ground truth for ``award winning soft chocolate chip cookies" and its corresponding retrieved recipes.}
  \label{retrieve1}
\vspace{-0.1in}
\end{figure*}

\begin{figure*}
    \centering
    \begin{subfigure}[b]{\textwidth}
        \centering
        \includegraphics[width=\linewidth, height=0.6\linewidth]{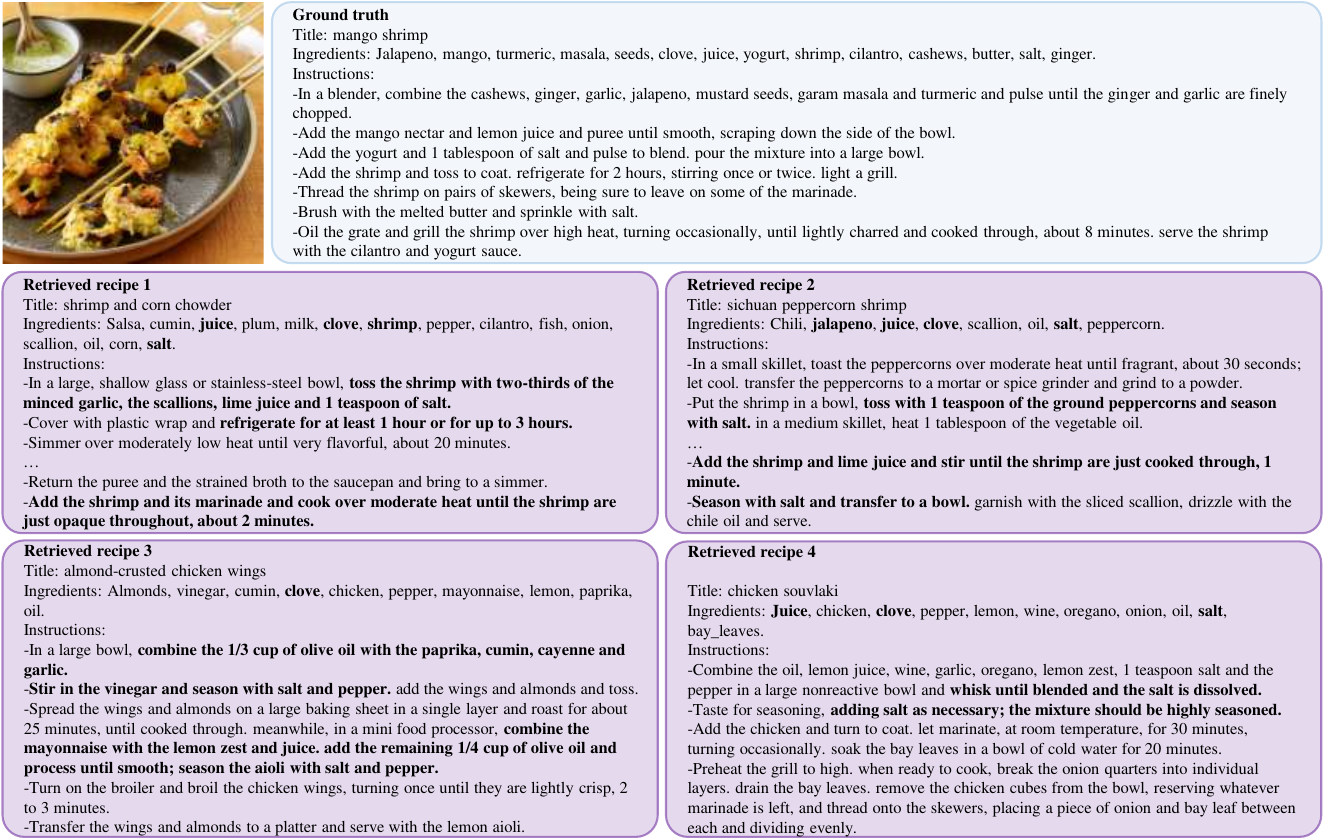}
        \captionsetup{skip=1pt}
        \caption{The example of the ground truth for ``mango shrimp" and its corresponding retrieved recipes.}
        \label{fig:a}
    \end{subfigure}

    \begin{subfigure}[b]{\textwidth}
        \centering
        \includegraphics[width=\linewidth, height=0.6\linewidth]{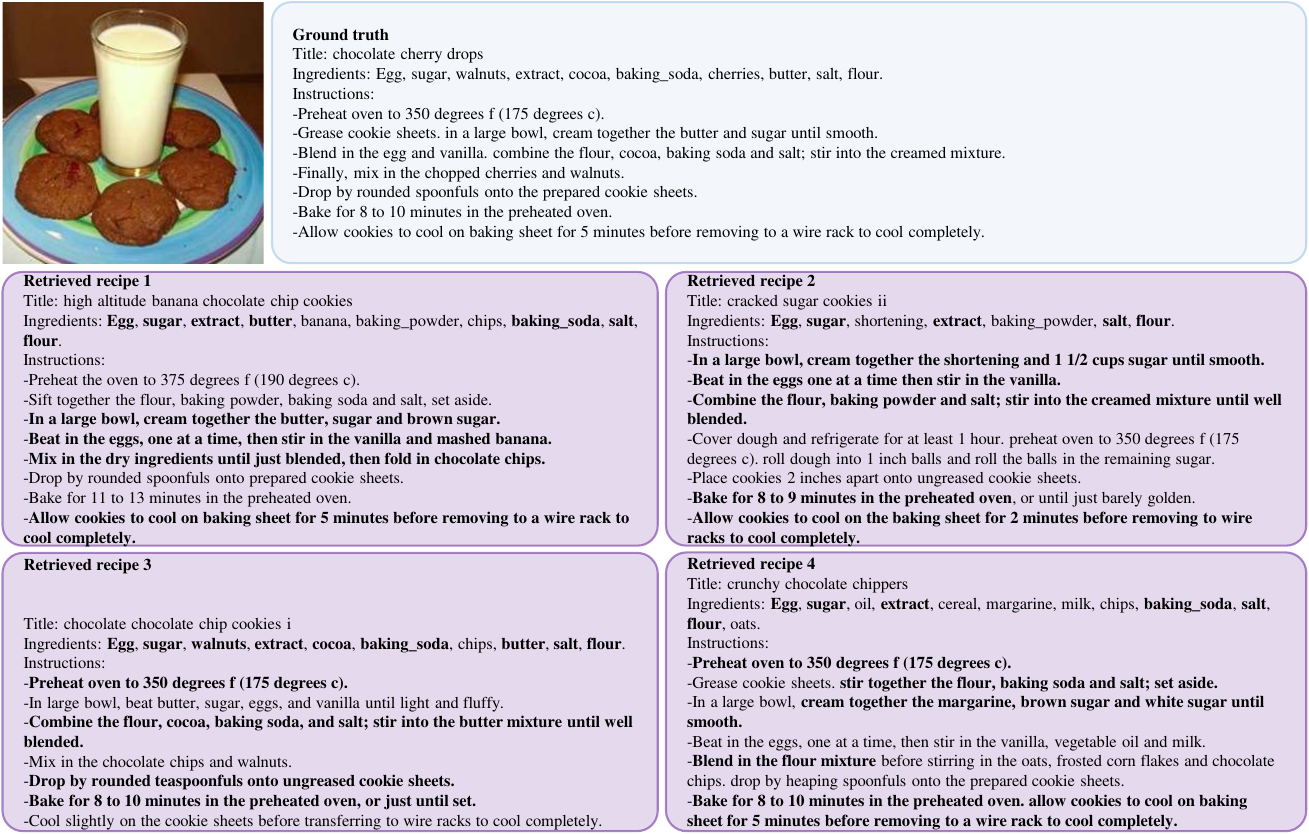}
        \captionsetup{skip=1pt}
        \caption{The example of the ground truth for ``chocolate cherry drops" and its corresponding retrieved recipes.}
        \label{fig:b}
    \vspace{-0.1in}
    \end{subfigure}
    \caption{Comparison between retrieval results and ground truth.}
    \label{retrieve}
\vspace{-0.1in}
\end{figure*}

Our SDRA method utilizes a variety of retrieved recipes to enhance model performance. Figure \ref{retrieve1}, and Figure \ref{retrieve} respectively illustrate some of the training data alongside their corresponding retrieved data. It can be seen that there are significant overlaps between the retrieved ingredients and the ground truth ingredients, which means the retrieved ingredients can provide additional information to enhance the model's diversity based on the original ingredients. For the retrieved instructions, they provide additional information based on many relevant contents to the ground truth. For example, in Figure \ref{retrieve1}, concerning the food ``award winning soft chocolate chip cookies," ``three out of four retrieved results precisely matched the preparation process for ``chocolate chip cookies." Steps like ``Preheat oven to 350 degrees f (175 degrees c)." and ``Blend in the dry ingredients, then fold in the chocolate chips." align with the ground truth instructions, as indicated by the BOLD parts in the diagram, ensuring that the additional information effectively provides content relevant to the original query. Additionally, steps from Retrieved recipe 4 like ``Allow cookies to cool for 1 minute on baking sheets before transferring to wire racks to cool completely." add detailed descriptions to the cookie-making process, enhancing the post-preparation flow and providing the model with more detailed and comprehensive information to complement and diversify the instructions found in the ground truth. Similarly, Figure \ref{retrieve} demonstrates that the retrieved recipes add useful contextual information related to the original queries to the Ground Truth. This guides the model to generate recipes more effectively, perfectly leveraging retrieval augmentation technology for enhanced predictions. This adds fundamental procedural details to the more complexly seasoned ground truth instruction, enabling the model to accurately generate responses to queries about the food image using the given context. 

\section{Ablation of Stochastic Diversified Retrieval Augmentation (SDRA)} 
As described in Section 4.3.1, to investigate whether the ingredients and instructions added before $Q_{instructions}$ in Recipe demonstration $R$ need to come from the same top retrieval results, i.e., having the same $K$ value, we compared experiments where ingredients and instructions were sourced from the same top $K$ retrieval results, specifically concatenating the top $K$ ingredients and top $K$ instructions in sequence before $Q_{instructions}$. 
Table \ref{ablation_independent} indicates that the model with independent ingredients and instructions performs better, as our method SDRA relies on a broader range of knowledge and diversified retrieval information settings.
\begin{table}[]
\centering
\caption{Ablation study of the independence of retrieved ingredients and retrieved instructions. ``SDRA(Matched)" refers to the situation where retrieved ingredients and instructions are matched, while ``SDRA(Independent)" indicates that retrieved ingredients and instructions are mutually independent.}
\label{ablation_independent}
\scriptsize
\begin{tabular}{cl|ccc}
\hline
\multicolumn{2}{c|}{Methods}          & BLEU  & SacreBLEU & ROUGE L \\ \hline
LLaVA-FT & \hspace{-1em}              & 28.32 & 5.88      & 38.18   \\ \hline
         & \hspace{-1em}+SDRA(Matched)     &28.83      &6.17           &38.30         \\ \hline
         & \hspace{-1em}+\textbf{SDRA(Independent)} &\textbf{29.23}      &\textbf{6.21}           &\textbf{38.43}         \\ \hline
\end{tabular}
\end{table}

\section{Ablation of Self-consistency Ensemble Voting}
To demonstrate the impact of self-consistent ensemble voting, Table \ref{ablation_vote} examines the results of our model when using cosine similarity, BLEU, SacreBLEU, and ROUGE L to calculate the mutual agreement among the generated recipe candidates. Note that $S=1$ refers to no voting adopted, which is the baseline for our model with Self-consistency Ensemble Voting when $S>1$. In other words, the input prompt is concatenated with the top 1 retrieved ingredients and instruction before $Q_{instructions}$ during inference.
To ensure the fairness of the experiments, these experiments start with the SDRA(top 50) model. The experiments with different voting groups indicate using the top $S$ retrieved instructions and ingredients for $S$ separate predictions as described in Section 3.3, and then selecting the score of the prediction with the highest score of mutual agreement as the final output. As $S$ is increased from 1 to 11, the voting results based on Cosine Similarity as the scoring metric show a steady improvement across all metrics. The results based on the other three scoring metrics also exhibit fluctuating improvements across various indicators, and after S=7, there is a trend of slower growth, no further increase, or even a decrease in some metrics.
\begin{table}[]
\centering
\caption{Ablation study of Self-consistency Ensemble Voting. Four scoring metrics are examined to evaluate the mutual agreement among generated recipes, including cosine similarity, BLEU, SacreBLEU, and ROUGE L. ``S" refers to the number of generated recipes for ensemble voting. ``Sum" is the sum of 3 evaluation metrics.}
\label{ablation_vote}
\scriptsize
\begin{tabular}{c|c|cccc}
\hline
Scoring metric                                                                & Number & BLEU           & SacreBLEU     & ROUGE L        & Sum            \\ \hline
\multirow{6}{*}{\begin{tabular}[c]{@{}c@{}}Cosine \\ Similarity\end{tabular}} & S=1    & 29.23          & 6.21          & 38.43          & 73.87          \\ \cline{2-6} 
                                                                              & S=3    & 29.68          & 6.31          & 38.68          & 74.67          \\ \cline{2-6} 
                                                                              & S=5    & 30.12          & 6.39          & 38.66          & 75.17          \\ \cline{2-6} 
                                                                              & S=7    & 30.07          & 6.41          & 38.84          & 75.32          \\ \cline{2-6} 
                                                                              & S=9    & 30.11          & 6.42          & 38.91          & 75.44          \\ \cline{2-6} 
                                                                              & S=11   & 30.11          & 6.42          & 38.93          & 75.47          \\ \hline
\multirow{6}{*}{BLEU}                                                         & S=1    & 29.23          & 6.21          & 38.43          & 73.87          \\ \cline{2-6} 
                                                                              & S=3    & 29.25          & 6.27          & 38.88          & 74.40          \\ \cline{2-6} 
                                                                              & S=5    & 29.47          & 6.35          & 38.93          & 74.75          \\ \cline{2-6} 
                                                                              & S=7    & 29.50          & 6.36          & 39.09          & 74.95          \\ \cline{2-6} 
                                                                              & S=9    & 29.53          & 6.37          & 39.09          & 74.99          \\ \cline{2-6} 
                                                                              & S=11   & 29.52          & 6.37          & \textbf{39.13} & 75.02          \\ \hline
\multirow{6}{*}{SacreBLEU}                                                    & S=1    & 29.23          & 6.21          & 38.43          & 73.87          \\ \cline{2-6} 
                                                                              & S=3    & 28.79          & 6.21          & 39.11          & 74.11          \\ \cline{2-6} 
                                                                              & S=5    & 29.30          & 6.30          & 38.83          & 74.43          \\ \cline{2-6} 
                                                                              & S=7    & 29.33          & 6.32          & 39.01          & 74.66          \\ \cline{2-6} 
                                                                              & S=9    & 29.33          & 6.32          & 39.10          & 74.75          \\ \cline{2-6} 
                                                                              & S=11   & 29.30          & 6.35          & 39.10          & 74.74          \\ \hline
\multirow{6}{*}{ROUGE L}                                                      & S=1    & 29.23          & 6.21          & 38.43          & 73.87          \\ \cline{2-6} 
                                                                              & S=3    & 32.21          & 6.62          & 36.99          & 75.82          \\ \cline{2-6} 
                                                                              & S=5    & 32.92          & 6.81          & 36.90          & 76.63          \\ \cline{2-6} 
                                                                              & S=7    & 33.08          & 6.82          & 36.80          & 76.70          \\ \cline{2-6} 
                                                                              & S=9    & 33.26          & 6.84          & 36.67          & 76.77          \\ \cline{2-6} 
                                                                              & S=11   & \textbf{33.53} & \textbf{6.88} & 36.60          & \textbf{77.01} \\ \hline
\end{tabular}
\end{table}
\section{More Qualitative Examples}
\begin{figure*}
  \centering
  \includegraphics[width=\linewidth]{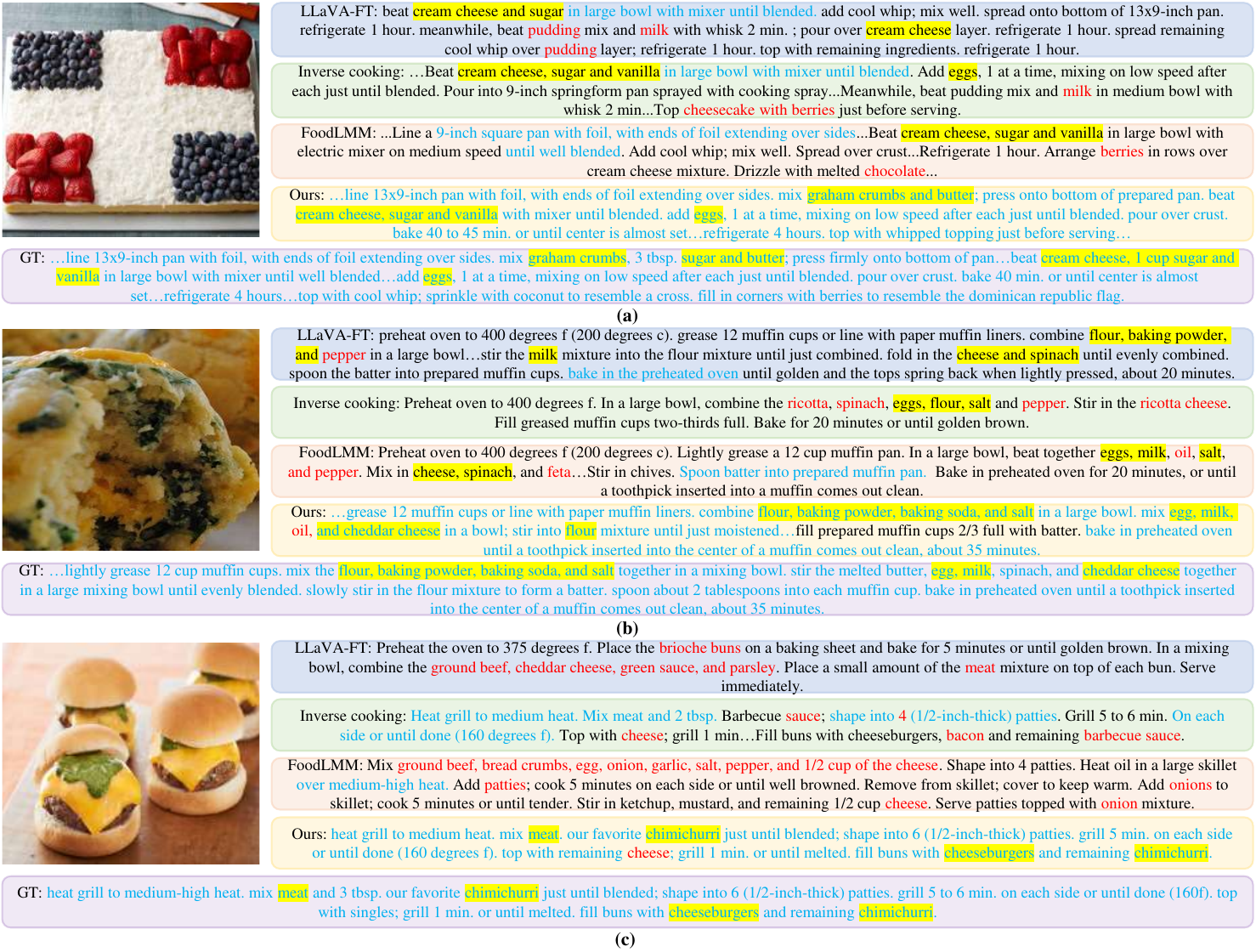}
  \caption{Additional qualitative results. The ingredients in generated recipes that overlap with ground truth (``GT") are highlighted in yellow, while details in the instructions that match the GT are shown in blue. Otherwise, the incorrect generation results are displayed in red. Best viewed in color.}
  \label{supp_data}
\end{figure*}

Figure \ref{supp_data} displays additional qualitative results. In these three cases, our predictions closely align with the ground truth ``GT", whereas other models—``LLaVA-FT", ``Inverse Cooking" \cite{Inverse_Cooking}, and ``FoodLLM" \cite{FoodLMM} —exhibited various hallucinations. For instance, in Figure \ref{supp_data} (a), ``LLaVA-FT" incorrectly predicted ``coconut" as ``pudding", ``Inverse Cooking" identified the food as ``cheesecake", and ``FoodLLM" mistakenly labeled blueberries as ``chocolate", though their overall instruction predictions were reasonably accurate. In Figure \ref{supp_data} (b), ``LLaVA-FT" and ``FoodLLM" made slight errors in ingredient prediction but were generally close to ``GT", however, their instruction predictions did not closely match ``GT", leading to lower test scores, whereas ``Inverse Cooking" provided overly simplistic and less detailed predictions. In Figure \ref{supp_data} (c), both ``LLaVA-FT" and ``FoodLLM" misidentified ``meat" as ``ground beef", affecting the accuracy of their instruction predictions, while ``Inverse Cooking" was more accurate, though it lacked the detail of our model, such as ``4 (1/2-inch-thick) patties".



\begin{figure*}
  \centering
  \includegraphics[width=\linewidth]{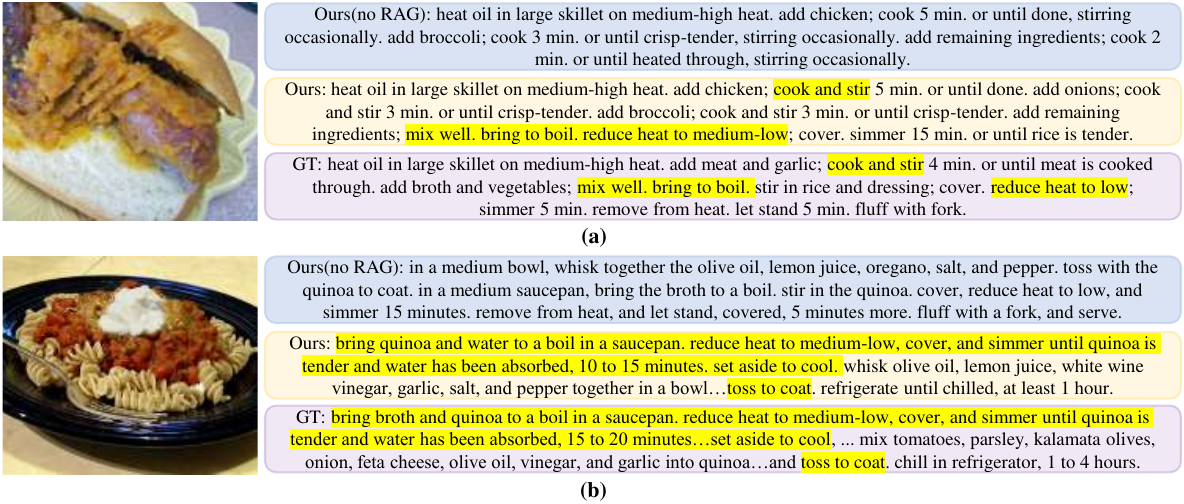}
  \caption{Qualitative results of whether the best model uses retrieval information for inference. The yellow highlights indicate the parts where our method provides more accurate predictions on the best model compared to the method without retrieval information.}
  \label{without_RAG}
\end{figure*}

Figure \ref{without_RAG} demonstrates the comparative results of our best model during the inference process, with and without the addition of retrieval information. It is evident that the predictions made without retrieval information are less accurate and comprehensive than those made with it. For instance, in Figure \ref{without_RAG}(a), our method precisely predicts ``cook and stir" instead of merely ``cook," in comparison to the results without added retrieval information. In Figure \ref{without_RAG}(b), our method's predictions almost perfectly match the ground truth, further underscoring the importance of incorporating retrieval information.

\begin{figure*}
  \centering
  \includegraphics[width=\linewidth]{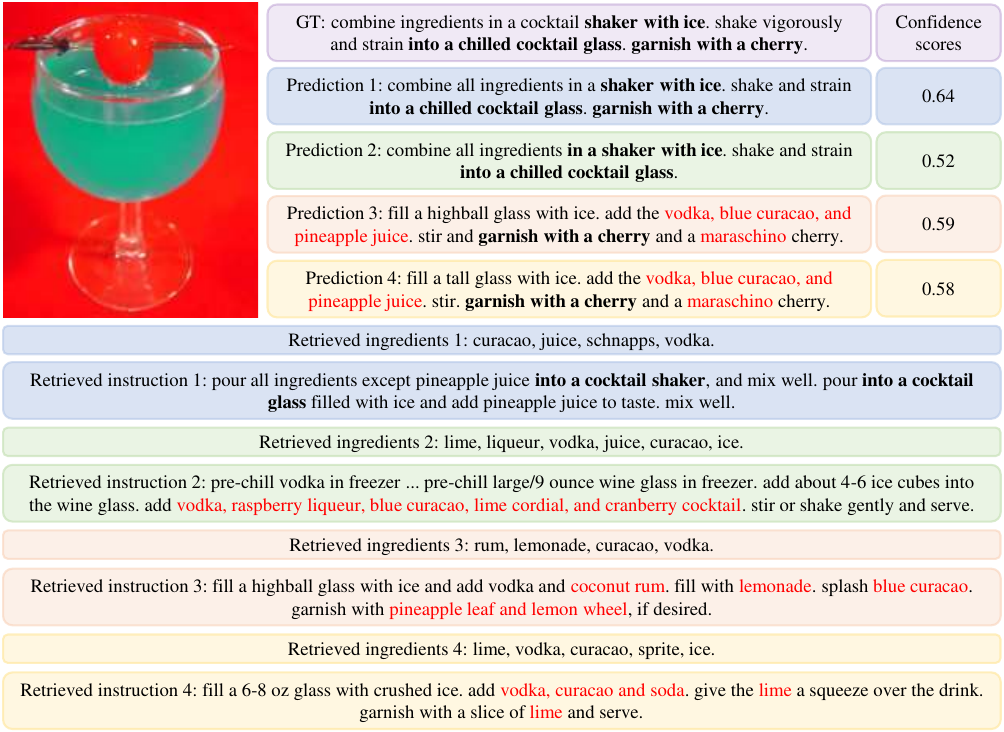}
  \caption{Qualitative results of Ablation of Self-consistency Ensemble Voting. \textbf{BOLD} indicates the parts that overlap with the ground truth, while red text highlights the hallucinated outputs. The retrieved ingredients and instructions correspond to each prediction, and the confidence score is calculated from a 4x4 cosine similarity matrix when S=4, with each row averaged to obtain the final score.}
  \label{case}
\end{figure*}

Figure \ref{case} shows the four generated recipe candidates, with prediction 1 being the output with the highest confidence. It overlaps with the ground truth in `shaker with ice' and `a chilled cocktail glass.' 
It can be observed that Prediction 3 and Prediction 4's `vodka, blue curacao, and pineapple juice' received supplementary information from their corresponding retrieved ingredients and instructions, such as `blue curacao' and `vodka, curacao, and soda'. However, the description in the retrieved information differs significantly from the ground truth, resulting in a lower confidence score for the predictions.
This demonstrates that the self-consistency voting mechanism effectively selects the best recipe.

\section{Consistency between the Generated Ingredients VS Retrieved Ingredients}
\begin{figure*}
  \centering
  \includegraphics[width=\linewidth]{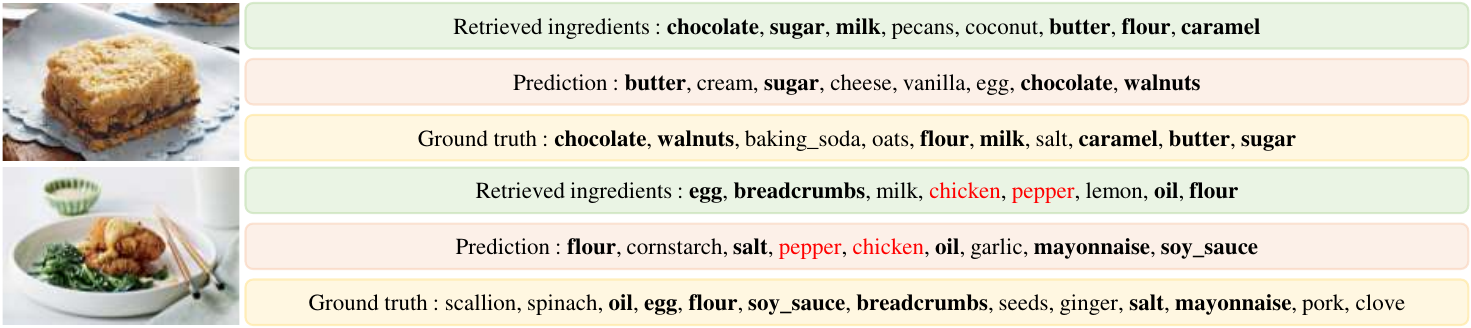}
  \caption{Retrieved and predicted ingredients. \textbf{BOLD} represents the overlap of ingredients between both Retrieved ingredients and Predicted ingredients with the ground truth ingredients. The ingredients in red font are the ones incorrectly predicted due to errors in the retrieved ingredients.}
  \label{111}
\end{figure*}

We compared the retrieved ingredients with the ingredients predicted by the model. As shown in Figure \ref{111}, in the first case, it can be observed that the model prediction partly overlaps with the ground truth. However, the retrieved ingredient list misled the model by introducing `cheese' and `vanilla', which are not part of the ground truth. The retrieved `walnuts' were beneficial, helping the model predict `walnuts' correctly. In the second case, the retrieved ingredients misled the prediction by introducing incorrect ingredients (`chicken', `pepper'), which were not part of the ground truth. Nonetheless, the model was able to predict other ingredients like `flour', `oil', `soy sauce', and `breadcrumbs' accurately. In both cases, the retrieved ingredients significantly influenced the model's prediction. When the retrieval stage introduced irrelevant ingredients (like chicken and pepper in the second case), these errors propagated into the prediction. However, the model was still able to predict a portion of the correct ingredients, particularly when they overlapped with the retrieved ones. The results underscore the importance of accurate ingredient retrieval, as incorrect retrievals can lead to faulty predictions, even if the model has the capacity to predict well when provided with accurate inputs.
\section{Ingredients Extracted from Instructions and Directly Predicted}
\begin{figure*}
  \centering
  \includegraphics[width=\linewidth]{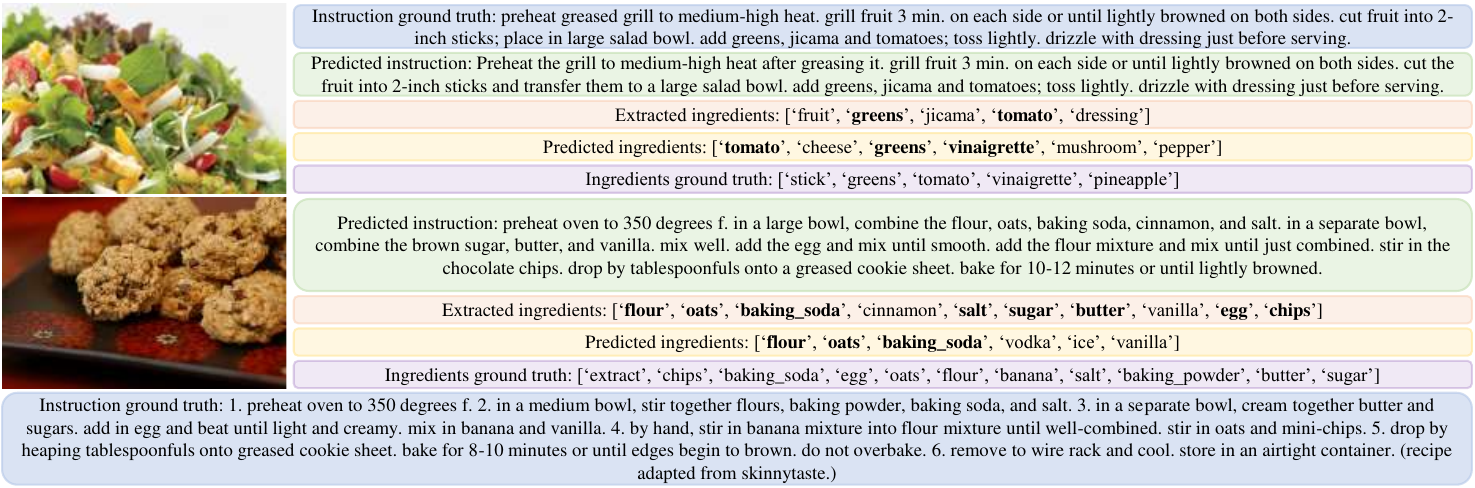}
  \caption{Predicted and ground truth ingredients, instructions, and ingredients extracted from the predicted instructions. \textbf{BOLD} represents the overlap of ingredients between both Extracted ingredients and Predicted ingredients with the ground truth ingredients.}
  \label{extract}
\end{figure*}

Figure \ref{extract} shows a comparison between ingredients extracted from generated instructions and those directly predicted. In the first case, ingredients like `greens' and `tomato', which are visually prominent in the image, are successfully predicted. However, `jicama' is detected in the instructions and appears in the instruction ground truth, but is missing from the directly predicted ingredients. This might be because the model, when generating instructions, learns common cooking pairings and automatically adds ingredients missed during ingredient prediction. Additionally, the retrieved instructions may have included `jicama'. It is worth noting that the ingredient `pineapple' in the extracted ingredients was identified as `fruit'. While this is correct in a broad or human sense, it would be considered incorrect when calculating classification metrics. In the second case, the extracted ingredients are almost entirely accurate, while the predicted ingredients are fewer but still quite accurate. This could be because the model uses more contextual information when generating instructions, which helps it infer ingredients that were not captured during standalone ingredient prediction. Overall, the ingredient predictions are fairly accurate, and there is a significant overlap between the extracted and predicted ingredients.

\begin{figure*}[h]
  \captionsetup{skip=5pt}
  \centering
  \includegraphics[width=\linewidth]{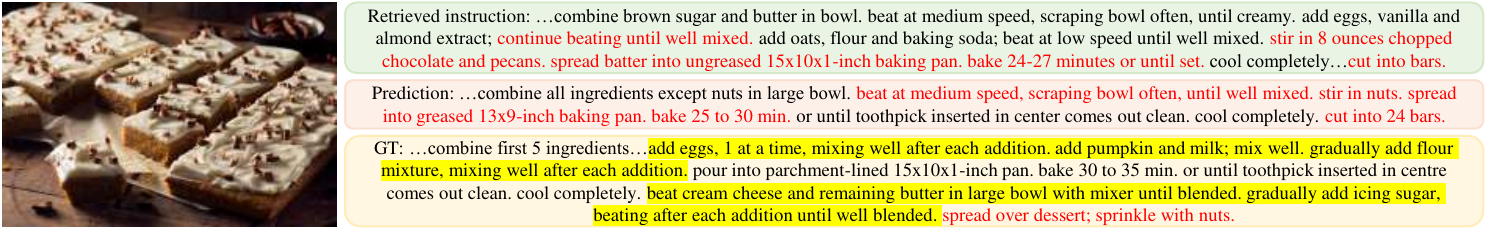}
  \caption{Retrieval failure examples. Red text shows prediction errors influenced by the retrieved instruction, and yellow highlights indicate missing content in the prediction.}
  \label{failure}
\vspace{-0.1in}
\end{figure*}

\begin{figure*}[h]
  \captionsetup{skip=5pt}
  \centering
  \includegraphics[width=\linewidth]{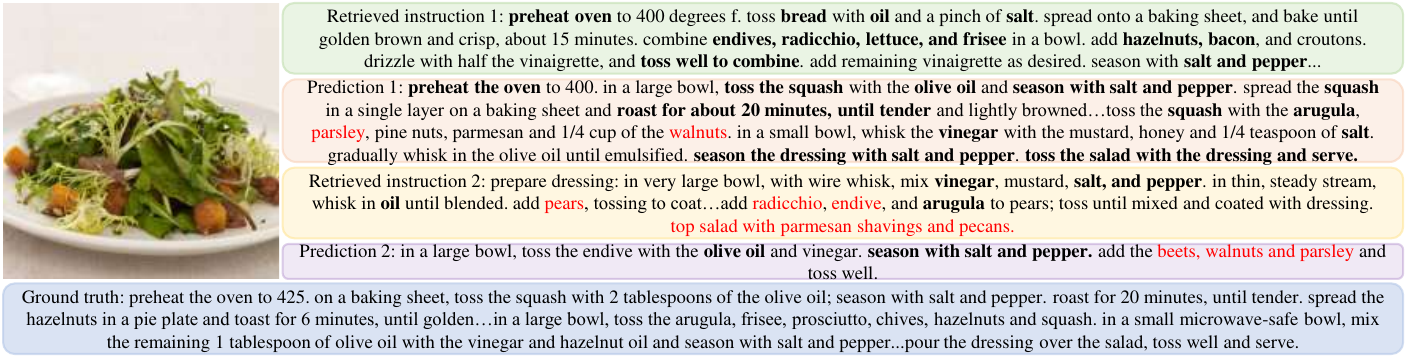}
  \caption{Two sets of retrieval results for ``butternut squash salad with hazelnuts" and their corresponding predictions. \textbf{BOLD} represents the parts that overlap with the ground truth, while red text indicates those that do not match the ground truth.}
  \label{noise}
\vspace{-0.1in}
\end{figure*}
\section{Retrieval Failure Case}
To explore retrieval failure cases, we analyzed some examples of unsuccessful retrievals. Figure \ref{failure} provides an example of a retrieval failure. The prediction is influenced by retrieved information, like "cut into bars," and mixing nuts into the batter early. In the ground truth, nuts are sprinkled on top at the end. Figure \ref{noise} presents two retrieved instructions for the salad and their respective predictions. It can be seen that the completeness of the preparation steps also influences the comprehensiveness of the prediction, and the predicted ingredients tend to overlap with the ingredients mentioned in the retrieved instructions. These results clearly show that when the retrieved instruction is more comprehensive and accurate, the resulting prediction will also be better. 
\section{Broader Ingredient Categories}
To avoid situations where ingredients in the prediction and ground truth are very similar but are counted as completely incorrect due to different terminology in the ingredients list—such as 'pasta' and 'spaghetti'—we designed an alternative penalty calculation method. We identified similar ingredients in the list and grouped them into 42 broader categories, as shown in Table \ref{42cate}. We apply a reduced penalty (0.5 weight) for ingredients within the same category and a full penalty (1 weight) for those across categories.
\begin{table}[h!]
\centering
\begin{tabular}{|l|p{12cm}|}
\hline
\textbf{Broader Category} & \textbf{Ingredients} \\ \hline
Pasta and Derivatives & spaghetti, pasta, penne, linguine, fettuccine, tortellini, ravioli, vermicelli, lasagna\_sheet, orecchiette, fusilli, conchiglie, cavatelli, spaghettini, manicotti \\ \hline
Cheese & cheese, medium\_cheddar, parmesan\_rind, parmigiano, gorgonzola, mascarpone, ricotta\_salata, queso\_fresco, feta \\ \hline
Meat & beef, chicken, pork, lamb, veal, bacon, ham, sausage, rib, tenderloin, fillets, steak, roast, duck, meatballs, sirloin, liver, tender\_quick \\ \hline
Fish and Seafood & tuna, salmon, shrimp, lobster, oyster, cod, scallops, sardines, anchovies, crabmeat, mussels, trout, mackerel, sole, haddock \\ \hline
Seasonings & salt, pepper, chili, oregano, basil, rosemary, thyme, parsley, cumin, coriander, paprika, mustard, dill, chives, cinnamon, nutmeg, allspice, marjoram, bay\_leaf, curry, saffron, tarragon, cardamom, ginger, garlic, horseradish, vanilla, extract \\ \hline
Vegetables & onion, celery, carrot, potato, zucchini, cucumber, lettuce, spinach, broccoli, cabbage, kale, squash, arugula, leek, fennel, asparagus, artichoke, beet, radish, tomato, eggplant, pumpkin \\ \hline
Fruits & apple, banana, berries, grapes, melon, cantaloupe, lemon, lime, orange, peach, pineapple, pear, mango, strawberry, kiwi, watermelon \\ \hline
Nuts and Seeds & peanuts, almonds, cashews, walnuts, pecans, hazelnuts, macadamias, sesame \\ \hline
Oils and Fats & oil, butter, margarine, shortening, lard, ghee \\ \hline
Dairy Products & milk, cream, yogurt, buttermilk, cheese \\ \hline
Flour and Grains & flour, cornmeal, oats, quinoa, barley, wheat, rice \\ \hline
Beans and Soy Products & lentils, chickpeas, kidney\_bean, soybeans, tofu, edamame, peas \\ \hline
Sweeteners and Sugars & sugar, honey, molasses, syrup, stevia, fructose \\ \hline
Baking Ingredients & baking\_soda, baking\_powder, yeast, vanilla, cocoa, gelatin, cornstarch \\ \hline
Sauces and Condiments & ketchup, mayonnaise, soy\_sauce, worcestershire\_sauce, teriyaki\_sauce, barbecue\_sauce, salad\_dressing, vinaigrette, gravy, mustard, hot\_sauce, ranch\_dressing, marinara\_sauce, pesto\_sauce, tartar\_sauce \\ \hline
Drinks & tea, coffee, wine, beer, lemonade, milk, brandy, rum, vodka, gin, cider, cola \\ \hline
Alcoholic Beverages & brandy, rum, vodka, gin, liqueur, champagne, vermouth, tequila \\ \hline
Processed and Seasoned Foods & bacon, sausage, hot\_dog, ham, salami, prosciutto \\ \hline
Processed Grain Foods & bread, cracker, chips, pie\_crust, pancake, waffle, biscuit \\ \hline
Dried Seafood & nori, kelp, bonito\_flakes, anchovy \\ \hline
Nut Candies and Pastries & peanuts, almonds, cashews, pecans, macadamias, hazelnuts, walnuts \\ \hline
Baking and Desserts & cake, cookie, brownie, muffin, pudding, pancake, waffle \\ \hline
Dry Goods and Grains & rice, oats, quinoa, barley, bulgur, millet, couscous \\ \hline
Broth and Seasoning Liquids & broth, stock, bouillon, gravy, miso \\ \hline
Tea and Coffee Beverages & coffee, espresso \\ \hline
Jams and Preserves & jam, jelly, marmalade, preserves \\ \hline
Candies and Sweets & candy, chocolate, fudge, caramel, marshmallow \\ \hline
Berries & strawberries, blueberries, raspberries, blackberries, cranberries \\ \hline
Tropical Fruits & mango, papaya, pineapple, banana, coconut \\ \hline
Citrus Fruits & orange, lemon, lime, grapefruit, tangerine \\ \hline
Leafy Vegetables & spinach, kale, lettuce, arugula \\ \hline
Root Vegetables & potato, carrot, beet, radish, turnip \\ \hline
Spices and Seasoning Powders & cumin, coriander, turmeric, paprika \\ \hline
Mushrooms and Fungi & mushroom, truffle, morel \\ \hline
Alcoholic Beverages & wine, beer, brandy, vodka, rum, gin, tequila \\ \hline
Legumes and Soy Products & kidney\_bean, chickpeas, lentils, soybeans, tofu \\ \hline
Soy Sauce and Asian Condiments & soy\_sauce, teriyaki\_sauce, hoisin\_sauce \\ \hline
Honey and Syrups & honey, molasses \\ \hline
Concentrated Sauces & ketchup, mustard, mayonnaise, barbecue\_sauce, salad\_dressing \\ \hline
Breads and Baked Goods & bread, baguette, bagel, muffin \\ \hline
Pasta Sauces & pesto\_sauce, marinara\_sauce, alfredo\_sauce, bolognese\_sauce \\ \hline
Grains and Cereals & rice, oats, quinoa, barley, bulgur \\ \hline
\end{tabular}
\caption{Broader categories and their respective ingredients.}
\label{42cate}
\end{table}

{\small
\bibliographystyle{ieee_fullname}
\bibliography{egbib}
}